\documentclass[10pt,twocolumn,letterpaper]{article}

\usepackage[final]{cvpr}
\usepackage{graphicx, amsmath, amssymb, xcolor, booktabs,colortbl}

\usepackage[font=small,labelfont=bf]{caption}
\usepackage{makecell, tabulary, pifont}

\definecolor{demphcolor}{RGB}{144, 144, 144}
\definecolor{mygray}{gray}{0.4}
\definecolor{LightCyan}{rgb}{0.9059,0.9961,1}
\newcommand{\demph}[1]{\textcolor{demphcolor}{#1}}

\newlength\savewidth

\newcommand{\tablestyle}[2]{\setlength{\tabcolsep}{#1}\renewcommand{\arraystretch}{#2}\centering\footnotesize}
\makeatletter\renewcommand\paragraph{\@startsection{paragraph}{4}{\z@}{.5em\@plus1ex\@minus.2ex}{-.5em}{\normalfont\normalsize\bfseries}}
\makeatother

\newcolumntype{C}[1]{>{\centering\arraybackslash}p{#1}}
\newcolumntype{R}[1]{>{\raggedleft\arraybackslash}p{#1}}
\newcolumntype{L}[1]{>{\raggedright\arraybackslash}p{#1}}

\newcommand\merlottitlefont[1]{{\usefont{T1}{cinzeldecorativebold}{m}{n}#1}}
\newcommand\merlotfont[1]{{\usefont{T1}{cinzeldecorative}{m}{n}#1}}
\newcommand{\modeltitle}{\merlottitlefont{VIOLET}}
\newcommand{\modelname}{\merlotfont{VIOLET}\xspace}

\usepackage[pagebackref,breaklinks,colorlinks]{hyperref}
\usepackage[capitalize]{cleveref}
\crefname{section}{Sec.}{Secs.}
\Crefname{section}{Section}{Sections}
\Crefname{table}{Table}{Tables}
\crefname{table}{Tab.}{Tabs.}

\usepackage{enumitem, graphicx, multirow, amsfonts, amsmath, verbatim, algorithm, algorithmicx, algpseudocode, booktabs, arydshln, times, epsfig, amssymb, pifont, capt-of, wrapfig, lipsum, hyperref}

\definecolor{asparagus}{rgb}{0.53, 0.66, 0.42}

\def\cvprPaperID{xxx}
\def\confName{CVPR}
\def\confYear{2022}

\begin{document}

\title{\modeltitle~\includegraphics[height=20pt]{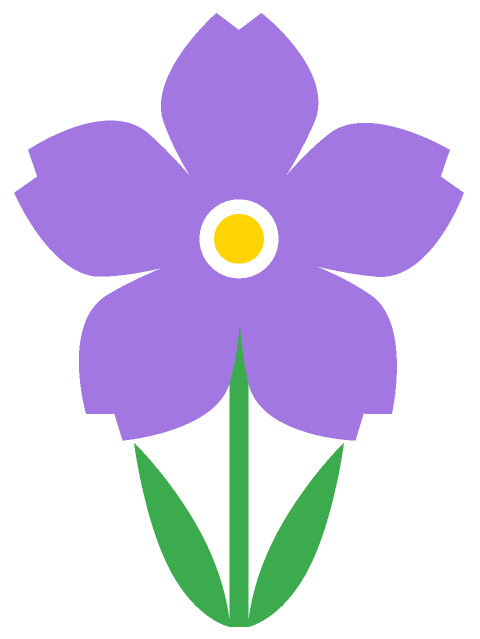}: End-to-End Video-Language Transformers with \\ Masked Visual-token Modeling}
\author{Tsu-Jui Fu$^\dagger$, Linjie Li$^\ddagger$, Zhe Gan$^\ddagger$, Kevin Lin$^\ddagger$, William Yang Wang$^\dagger$, Lijuan Wang$^\ddagger$, Zicheng Liu$^\ddagger$\\$^\dagger$UC Santa Barbara~~$^\ddagger$Microsoft\\{\tt \small \{tsu-juifu, william\}@cs.ucsb.edu}\\{\tt \small \{lindsey.li, zhe.gan, keli, lijuanw, zliu\}@microsoft.com}}
\maketitle

\begin{abstract}
A great challenge in video-language (VidL) modeling lies in the disconnection between fixed video representations extracted from image/video understanding models and downstream VidL data. Recent studies try to mitigate this disconnection via end-to-end training. To make it computationally feasible, prior works tend to ``imagify" video inputs, \textit{i.e.}, a handful of sparsely sampled frames are fed into a 2D CNN, followed by a simple mean-pooling or concatenation to obtain the overall video representations. Although achieving promising results, such simple approaches may lose temporal information that is essential for performing downstream VidL tasks. In this work, we present \modelname, a fully end-to-end VIdeO-LanguagE Transformer, which adopts a video transformer to explicitly model the temporal dynamics of video inputs. Further, unlike previous studies that found pre-training tasks on video inputs (\textit{e.g.}, masked frame modeling) not very effective, we design a new pre-training task, Masked Visual-token Modeling (MVM), for better video modeling. Specifically, the original video frame patches are ``tokenized'' into discrete visual tokens, and the goal is to recover the original visual tokens based on the masked patches. Comprehensive analysis demonstrates the effectiveness of both explicit temporal modeling via video transformer and MVM. As a result, \modelname achieves new state-of-the-art performance on 5 video question answering tasks and 4 text-to-video retrieval tasks.\footnote{Code is available at \url{https://github.com/tsujuifu/pytorch_violet}} 
\end{abstract}

\vspace{-2ex}
\section{Introduction}
Humans are born to perceive this world from multiple modalities such as vision, sound, and touch. Video, containing multiple modalities in nature, has been used as an epitome to test how AI systems perceive. Video-language (VidL) research aims at extending this ability to convey perception via language. Popular VidL tasks were introduced, such as text-to-video retrieval~\cite{xu2016msrvtt,krishna2017dense-caption,rohrbach2015lsmdc,lei2020tvr,li2020hero}, video question answering~\cite{jang2017tgif-qa,xu2017msrvtt-qa,lei2018tvqa,lei2020tvqa+}, text-based video moment retrieval~\cite{hendricks2017local-moment,gao2017tall,krishna2017dense-caption,lei2020tvr}, and video captioning~\cite{wang2019vatex,zhou2018youcook2,xu2016msrvtt,rohrbach2015lsmdc}. 

\begin{figure}[t]
\centering
    \includegraphics[width=\linewidth]{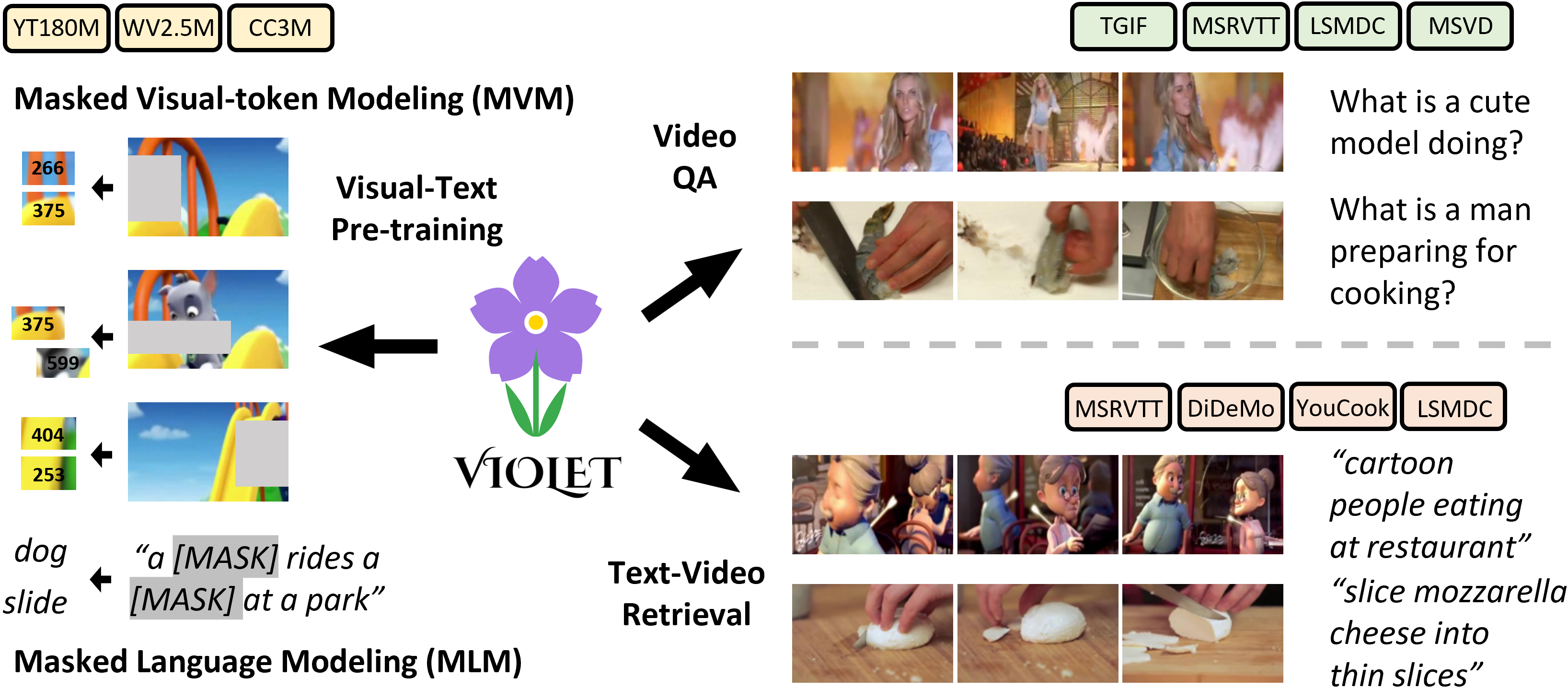}
    \vspace{-4ex}
    \caption{End-to-end \textbf{V}\textbf{I}de\textbf{O}-\textbf{L}anguag\textbf{E} \textbf{T}ransformer (\modeltitle). \modelname~performs large-scale visual-text pre-training and can be applied to various video question answering and text-to-video retrieval tasks.}
    \vspace{-2ex}
    \label{fig:intro}
\end{figure}

Previous works~\cite{jang2017tgif-qa,le2020hcr-vqa,zhu2020act-bert,miech2019howto100m,li2020hero,xu2021video-clip} attempt cross-modal fusion over dense video features and text features to tackle VidL tasks, but suffer from domain disconnection due to offline feature extraction~\cite{lei2021clip-bert,miech2019howto100m}. To address this issue, ClipBERT~\cite{lei2021clip-bert} proposes to ``imagify'' the dense video frame inputs. First, it adopts a sparse sampling strategy to employ only a handful of frames from the entire video for efficient end-to-end training. Second, the overall video representations are obtained through mean-pooling a sequence of frame features, individually computed by a 2D Convolutional Network. Although obtaining promising results, the brash mean pooling over individual frame features forfeits the crucial temporal information in video. To improve temporal modeling, recent works~\cite{zellers2021merlot,wu2021ha-net} concatenate all sparse-sampled frame features in chronological order, and directly enforce VidL learning  along with the text inputs. However, these methods still treat video frames as static images, and rely heavily on cross-modal fusion module to capture both temporal dynamics in videos and the alignment between visual and textual elements simultaneously.

We propose fully end-to-end VIdeO-LanguagE Transformer (\modelname) to enhance video modeling for better VidL modeling from two perspectives: ($i$) model architecture, and ($ii$) pre-training task design. In terms of \emph{model architecture}, instead of naive mean pooling or concatenation over a sequence of individual frame features, \modelname contains Video Swin Transformer that models video temporal explicitly for VidL learning~\cite{bertasius2021timesformer,liu2021video-swin}. Since the self-attention over spatial-temporal locality allows modeling variable sequence lengths, our video transformer support flexible VidL learning from both videos and static images.

In terms of \emph{pre-training tasks}, though the direct adoption of Masked Language Modeling~\cite{devlin2019bert} has proven effective in pre-training vision-language models, the attempt on similar masked modeling on vision inputs is not as successful. For example, Masked Region Modeling (MRM)~\cite{chen2020uniter} or Masked Frame Modeling (MFM)~\cite{li2020hero} aim to recover masked image regions or video frames. Despite of the different variants of MRM/MFM that model object category or distilled region/frame features, it suffers from imperfect patch labels, excessive feature dimensions, rendering unsatisfactory performance~\cite{chen2020uniter,li2020hero}. Recent VidL works~\cite{lei2021clip-bert,zellers2021merlot,bain2021frozen} even completely discard such pre-training tasks due to limited performance improvements.

To promote better video representations for VidL learning, we present a new pre-training task: Masked Visual-token Modeling (MVM), as shown in the left of Fig.~\ref{fig:intro}. By using the pre-trained discrete VAE~\cite{oord2017vq-vae} from DALL-E \cite{ramesh2021dalle}, we ``tokenize'' the video frames into discrete visual tokens, which can be used to reconstruct the original video frames. During pre-training, we mask out some proportions of the video input along both spatial and temporal dimensions, and the model learns to recover the discrete visual tokens of these masked patches. MVM improves over previous MRM/MFM in two ways: ($i$) MVM predicts over a discrete space, which avoids falling ill with the similar training issues of excessive feature dimensions as in~\cite{chen2020uniter,li2020hero}; ($ii$) MVM is based on latent visual tokens obtained from a self-reconstruction training procedure, instead of distilling from a well-supervised visual backbone. Our comprehensive comparison shows that MVM enhances the model's capability to better understand video scenes and in turn benefits downstream VidL tasks.

In summary, our contributions are four-fold. ($i$) We present \modelname, a fully end-to-end transformer to model the spatial-temporal dynamic in videos for VidL learning. ($ii$) We propose a new pre-training task, Masked Visual-token Modeling, which recovers the masked video frame patches into a discrete visual token space. ($iii$) \modelname achieves state-of-the-art results on 4 text-to-video retrieval and 5 out of 8 video question answering tasks. ($iv$) Comprehensive ablation studies demonstrate the necessity of temporal video modeling and the effectiveness of MVM across different VidL pre-training settings.

\begin{figure*}[t]
\centering
    \includegraphics[width=\linewidth]{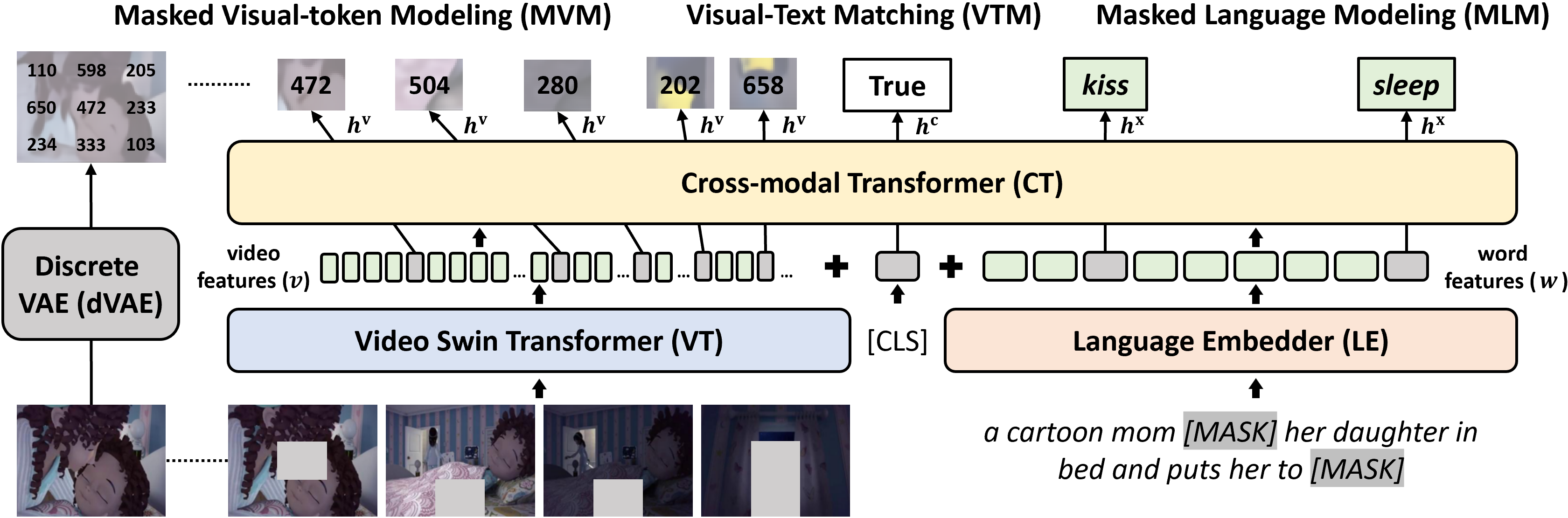}
    \vspace{-4ex}
    \caption{Overview of the proposed end-to-end \textbf{VI}de\textbf{O}-\textbf{L}anguag\textbf{E} \textbf{T}ransformer (\modeltitle), with Video Swin Transformer, Language Embedder, and Cross-modal Transformer. \modelname adopts Discrete VAE to extract discrete visual tokens to perform Masked Visual-token Modeling along with Visual-Text Matching and Masked Language Modeling during large-scale visual-text pre-training.}
    \vspace{-2ex}
    \label{fig:violet}
\end{figure*}

\section{Related Work}
\vspace{0.5ex} \noindent\textbf{Video-Language Understanding.}
Joint video-language (VidL) understanding~\cite{li2021value,liu2020collaborative-expert,jiang2020dac,le2020hcr-vqa,gabeur2020mmt,patrick2021support-set} aims at interpreting the physical world via both vision and text perception.  Researchers have explored such capability on VidL tasks including text-based video retrieval~\cite{xu2016msrvtt,krishna2017dense-caption,rohrbach2015lsmdc,lei2020tvr,li2020hero}, video question answering~\cite{jang2017tgif-qa,xu2017msrvtt-qa,lei2018tvqa,lei2020tvqa+}, moment retrieval~\cite{hendricks2017local-moment,gao2017tall,krishna2017dense-caption,lei2020tvr}, and video captioning~\cite{wang2019vatex,zhou2018youcook2,xu2016msrvtt,rohrbach2015lsmdc}. Prior arts before the large-scale pre-training era~\cite{ gao2018motion,zhang2018video-text,lei2021qvhighlights, fan2019heterogeneous,le2020hcr-vqa,lei2020tvqa+} leverage offline extracted video features~\cite{kay2017kinetics,wang2016temporal, carreira2017quo, xie2018rethinking,feichtenhofer2019slowfast,deng2009imagenet,he2016resnet,krishna2017vg,anderson2018bottom}. Later on, VidL pre-trained models~\cite{sun2019videobert,zhu2020act-bert,li2020hero,miech2019howto100m} built on the above pre-extracted features have shown promising results. To enhance the performance, there have been parallel interests in bringing in more modalities from raw video inputs~\cite{gabeur2020mmt,rouditchenko2021avlnet,liu2021hit} and end-to-end training till the raw pixels~\cite{miech2020end,lei2021clip-bert,zellers2021merlot,bain2021frozen}, both aiming to elevate video representations for VidL modeling.

Our work further explores the second direction for general VidL understanding, in contrast to~\cite{bain2021frozen} focusing on text-to-video retrieval only. Instead of encoding each video frame individually as static image and applying simple mean-pooling or concatenation along the temporal dimension~\cite{lei2021clip-bert,zellers2021merlot}, we demonstrate the necessity of temporal modeling by our video transformer over the input video frames, even when they are sparsely sampled~\cite{lei2021clip-bert}.

\vspace{0.5ex} \noindent\textbf{Masked Visual Modeling.}
Aligned with the success of transformer-based~\cite{vaswani2017attention} language pre-training models~\cite{liu2019roberta,yang2019xlnet,raffel2020exploring,lan2019albert,clark2020electra}, image-text pre-training~\cite{kim2021vilt,desai2021virtex,lu202012in1,su2020vlbert,zhou2020unified-vl,li2020unicoder-vl,li2019visual-bert,gan2020large,sun2021lightningdot,zhou2021uc2} and video-text pre-training~\cite{kim2021sspt-crl-vqa,yang2020bert-vqa,yang2021just-ask} have shown promising results on diverse vision-language (VL) tasks. Popular VL pre-training tasks include Visual-Text Matching and Masked Language Modeling, which are directly adapted from language pre-training~\cite{devlin2019bert}. Similar masked modeling on visual inputs~\cite{chen2020uniter,li2020hero} has also been introduced to VL pre-training, but are not as useful.  We propose Masked Visual-token Modeling (MVM), adopting the latent codes of discrete VAE~\cite{ramesh2021dalle,rolfe2017discrete-vae,oord2017vq-vae} as the reconstruction target for masked patches, which eases the auto-encoding prediction and can lead to a more significant improvement. 

Among the literature, BEiT~\cite{bao2021beit} and VIMPAC~\cite{tan2021vimpac} are two relevant studies of masked visual modeling for image classification~\cite{deng2009imagenet} and action recognition~\cite{kay2017kinetics}. Specifically, BEiT~\cite{bao2021beit} proposes a BERT-like pre-training strategy to recover the original visual tokens from some masked image patches. Our MVM takes inspiration from BEiT, but extends to more complex video inputs with an additional temporal dimension for VidL modeling. To prevent the model from taking shortcuts in recovering visual tokens from its spatial or temporal neighbors, we further introduce a combination of blockwise masking and attended masking. VIMPAC~\cite{tan2021vimpac} takes a step further to completely remove the raw pixel inputs from the training procedure. It employs visual tokens as the discrete representation of video inputs and applies a mask-then-predict pre-training task. The removal of raw pixel inputs renders a weaker baseline on popular action recognition tasks. In our work, we leverage visual tokens as prediction targets for MVM, instead of replacing the raw video frame patches.

\section{\modeltitle}
\subsection{Model Architecture}\label{sec:model}
Fig.~\ref{fig:violet} illustrates the overall architecture of our end-to-end video-language transformer (\modelname). \modelname contains 3 components: Video Swin Transformer (VT), Language Embedder (LE), and Cross-modal Transformer (CT).  \modelname takes video $\mathcal{V}$ and sentence $\mathcal{X}$ as inputs. Sparse-sampled frames $\{f_1, f_2, ...\}$ from $\mathcal{V}$ are first processed by VT to compute video features $v=\{v_1, v_2, ...\}$. LE extracts the word embeddings $w=\{w_1, w_2, ...\}$ for each word token $\{x_1, x_2, ...\}$ in $\mathcal{X}$. Then CT performs cross-modal fusion on top of $v$ and $w$ to produce joint video-language (VidL) representations $h$ for pre-training and downstream finetuning. We explain each component in detail below. 

\vspace{0.5ex} \noindent\textbf{Video Swin Transformer (VT).}
Instead of mean pooling or concatenating individual frame features, we adopt Video Swin Transformer~\cite{liu2021video-swin} (VT) to model $T$ sparse-sampled frames $\{f_t\}_{t=1}^T$  along both spatial and temporal dimensions as video features $\{v_t\}_{t=1}^T$. VT first splits each frame as non-overlapping $H \times W$  patches~\cite{dosovitskiy2021vit} and adopts a linear projection layer to obtain the preliminary video patch embeddings $u \in \mathbb{R}^{T \times H \times W \times d}$:
\begin{equation}
    u_{t} = \text{LinearProj}(f_{t}).
\end{equation}
The multi-layer 3D-shifted window~\cite{liu2021video-swin} then considers different levels of spatial-temporal attention over these video patch embeddings. We add learnable positional embedding $p^\text{v}$ to $u$, including spatial $p^\text{s} \in \mathbb{R}^{H \times W \times d}$ and temporal ordering $p^\text{t} \in \mathbb{R}^{T \times d}$, and extracts the video features $v$:
\begin{equation}
\begin{split}
    p^\text{v}_t &= p^\text{s}+p^\text{t}_t, \\
    v &= \text{VT}(\{u_t+p_t^\text{v}\}_{t=1}^T).
\end{split}
\end{equation}
All patches from the $t^\text{th}$ frame shares the same $p^\text{t}_t$ and all patches with the same spatial position are given the same $p^\text{s}$. In particular, each 3D window is in the size of $T' \times M \times M$ and considers video temporal across $T'$ consecutive frames. By adopting 3D windows upon blocks of video patches, VT can model image spatial and video temporal simultaneously through the self-attended computation procedure. Note that we make a slight modification to remove the temporal down-sampling from the original Video Swin Transformer and ensure the same temporal dimension as the input video for Masked Visual-token Modeling during pre-training (Sec.~\ref{sec:pre-training-task}).

VT enforces spatial-temporal modeling via  3D-shifted window to compute the initial video representations for VidL modeling. We demonstrate the advantages of VT over simple mean-pooling or concatenation of ``imagified'' frame representations under different VidL pre-training settings in Sec.~\ref{sec:video-encoding}. In addition, as VT encodes video frame patches through a fully self-attended computation, it can support a variable length of visual inputs. This video encoding enables \modelname to carry out static images (\textit{i.e.}, $T$ = 1). We discuss the flexibility of pre-training \modelname on both large-scale image-text data and video-text data in Sec.~\ref{sec:pretraining-data}. 

\vspace{0.5ex} \noindent\textbf{Language Embedder (LE).}
For a language input $\mathcal{X}$, we follow WordPiece~\cite{wu2016word-piece} and tokenize it into word tokens $\{x_i\}_{i=1}^L$, where $L$ is the number of tokens in $\mathcal{X}$. LE embeds the discrete word token $x_i$ into high-dimensional word representation $w_i \in \mathbb{R}^d$ :
\begin{equation}
    \{w_i\}_{i=1}^L = \text{LE}(\{x_i\}_{i=1}^L).    
\end{equation}

\vspace{0.5ex} \noindent\textbf{Cross-modal Transformer (CT).}
Given video features $v$ and word features $w$, CT performs cross-modal fusion over all $\{v_i\}_{i=1}^{T}$ and $\{w_i\}_{i=1}^L$ for joint VidL learning. We add different positional embeddings $p^\text{v}$ or $p^\text{x}$ to  video features $v$ or word features $w$, to incorporate sequence ordering and distinguish between the two modalities. In particular, we reuse $p^\text{v}$ from VT, containing both spatial position and temporal ordering information. We concatenate the video and text representations after position embedding as the input sequence to CT. In addition, a special \texttt{[CLS]} token is added to compute the global VidL representation, used in pre-training and downstream finetuning. The joint VidL features $h \in \mathbb{R}^{(T+1+L) \times d}$ are computed as:
\begin{equation}
\begin{split}
    h &= \text{CT}([v+p^\text{v}, \texttt{[CLS]}, w+p^\text{x}]), \\
    [h^\text{v}, &~h^\text{c}, h^\text{x}] = h,
\end{split}
\end{equation}

\subsection{Pre-training Tasks.} \label{sec:pre-training-task}
To benefit from large-scale data~\cite{zellers2021merlot,bain2021frozen,sharma2018cc}, we incorporate three pre-training tasks, including our proposed Masked Visual-token Modeling. Masked Language Modeling~\cite{devlin2019bert,chen2020uniter,li2020hero} predicts the masked word tokens to improve language reasoning with the aid of visual perception. Masked Visual-token Modeling recovers the masked video patches to enhance the video scene understanding. Visual-Text Matching~\cite{chen2020uniter,lei2021clip-bert,bain2021frozen} learns the alignments between video and text modality, improving the cross-modal fusion.

\vspace{0.5ex} \noindent\textbf{Masked Language Modeling (MLM).}
In MLM, we randomly mask out some word tokens with a probability of 15\%.\footnote{Following BERT~\cite{devlin2019bert}, We replace 80\% of masked word tokens as the \texttt{[MASK]} token, 10\% as a random token, and 10\% as its original token.} The goal is to recover these masked tokens $x$ from the joint VidL features $h$ modeled by Cross-modal Transformer (CT). Specifically, the corresponding $h^\text{x}$ for these masked tokens are fed in a fully-connected (FC) layer ($\text{FC}^\text{MLM}$) and projected to the discrete word token space for classification:
\begin{equation}
\begin{split}
    x'_i &= \text{FC}_\text{MLM}(h^\text{x}_i), \\
    \mathcal{L}_\text{MLM} &= - \mathbb{E}~[\frac{1}{|\mathcal{M}^\text{MLM}|} \sum\nolimits_{i \in \mathcal{M}^\text{MLM}} \log P(x_i~|~x'_i)],
\end{split}
\end{equation}
where $\mathcal{M}^\text{MLM}$ denotes the index set of masked word tokens. 

\vspace{0.5ex} \noindent\textbf{Visual-Text Matching (VTM).}
VTM enhances the cross-modal fusion via modeling the alignments between visual and textual inputs. At each training step, we randomly replace the corresponding text $\mathcal{X}_\text{pos}$ for a given video $\mathcal{V}$ with the text description $\mathcal{X}_\text{neg}$ from a different video in the same batch. Both the positive pair $(\mathcal{V}, \mathcal{X}_\text{pos})$ and negative pair $(\mathcal{V}, \mathcal{X}_\text{neg})$ are modeled by CT, and VTM is to tell them apart from the global VidL representation $h^\text{c}$ of the \texttt{[CLS]} token. In particular, $h^\text{c}$ will be processed by a FC layer ($\text{FC}^\text{VTM}$) to perform binary classification:
\begin{equation}
\begin{split}
    b_\text{pos} &= \text{FC}^\text{VTM}(h^\text{c}_\text{pos}), b_\text{neg} = \text{FC}^\text{VTM}(h^\text{c}_\text{neg}), \\
    \mathcal{L}_\text{VTM} &= - \mathbb{E} [\log(b_\text{pos}) + \log(1-b_\text{neg})],
\end{split}
\end{equation}
where $h^\text{c}_\text{pos}$ or $h^\text{c}_\text{neg}$ is $h^\text{c}$ of positive or negative pairs. 

\vspace{0.5ex} \noindent\textbf{Masked Visual-token Modeling (MVM).}
Previous Masked Region Modeling (MRM)~\cite{chen2020uniter} and Masked Frame Modeling (MFM)~\cite{li2020hero} extends MLM to visual inputs but sometimes leads to unsatisfactory performance. Different from MRM and MFM, which rely on distilled visual categories or features from a well-supervised visual backbone~\cite{anderson2018bottom,feichtenhofer2019slowfast}, we present Masked Visual-token Modeling (MVM) to perform masked visual modeling in a self-reconstruction scenario. We consider the discrete variational autoencoder (dVAE)~\cite{oord2017vq-vae,ramesh2021dalle} to quantize video inputs into masked prediction targets. dVAE is learned to tokenize images into discrete visual tokens $q$ from a finite vocabulary and then reconstruct the original visual scene based on $q$, where $q$ should have a one-to-one correspondence with the input image patches spatially. We first adopt dVAE to tokenize the $t^\text{th}$ video frame $f_t$ into $q_t$:
\begin{equation}
    q_t = \text{dVAE}(f_t).
\end{equation}
Similar to MLM, we mask out some video patches by replacing the pixel values with all zeros. MVM aims at recovering the visual tokens $q$ of those masked video patches $v$ from the corresponding joint VidL features $h^\text{v}$. $h^\text{v}$ is fed into a FC layer ($\text{FC}^\text{MVM}$) and projected to the discrete visual token space for classification:
\begin{align}
    q'_{t, i} &= \text{FC}^\text{MVM}(h^\text{v}_{t, i}), \\
    \mathcal{L}_\text{MVM} &= - \mathbb{E}~[\sum_{t=1}^T \frac{1}{|\mathcal{M}^\text{MVM}_t|} \sum\nolimits_{i \in \mathcal{M}^\text{MVM}_t} \log P(q_{t, i}~|~q'_{t, i})], \nonumber
\end{align}
where $\mathcal{M}^{\text{MVM}}_t$ is the index set of masked video patches for the $t^\text{th}$ frame. Using discrete visual tokens as masked prediction targets has two main advantages: ($i$) The finite vocabulary size of these discrete visual tokens eases the learning of MVM, avoid the previous difficulty in model training with MRM/MFM from imperfect patch categories or excessive feature dimensions; ($ii$) MVM does not require a well-supervised visual backbone to distill the masking labels. The latent visual tokens can be learned in a self-supervised way without human annotations.

\subsection{Masking Strategy of MLM and MVM}\label{sec:masking}

\begin{figure}[t]
\centering
    \includegraphics[width=.9\linewidth]{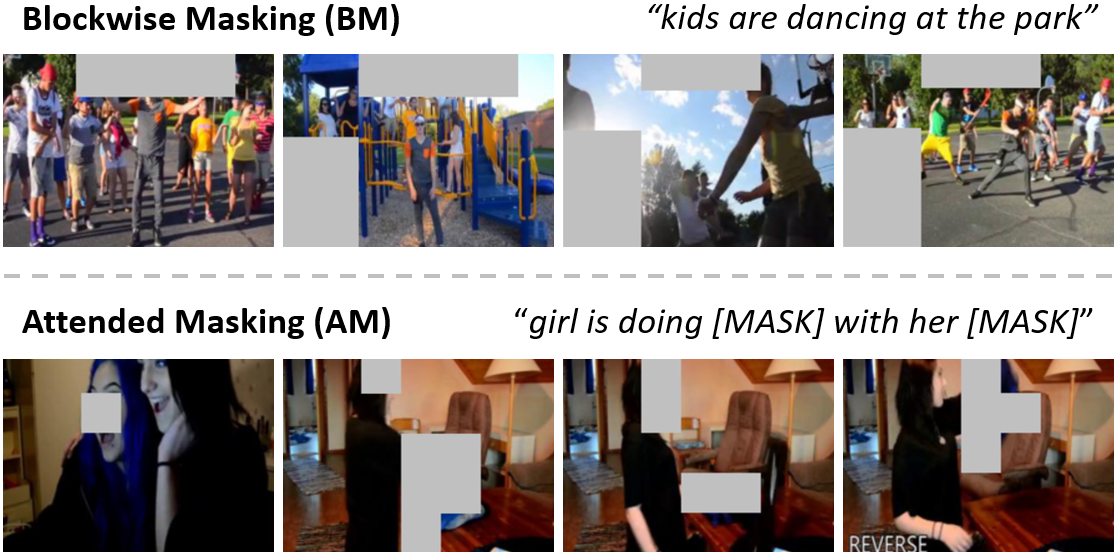}
    \vspace{-2ex}
    \caption{Masking Strategy of MLM and MVM, including \textbf{Blockwise Masking (BM)} and \textbf{Attended Masking (AM)}.}
    \vspace{-2ex}
    \label{fig:masking}
\end{figure}

We introduce a combination of Blockwise Masking and Attended Masking to amplify the effectiveness of MLM and MVM, as shown in Fig.~\ref{fig:masking}. 

\vspace{0.5ex} \noindent\textbf{Blockwise Masking (BM).}
Video usually presents analogous visual patterns in spatial-temporal neighbors (\textit{i.e.}, nearby patches within current frame or neighboring frames). While these neighbors make the masked video patches easy to recover, they may lead to spurious success in MVM evaluation. To make MVM more challenging, we adopt Blockwise Masking~\cite{tan2021vimpac,bao2021beit} that masks blocks of video patches along spatial-temporal dimension rather than independently masking randomly sampled patches for each frame. Specifically, we randomly sample an $(H', W', T')$ as a masking block, where all $H'\times W'$ visual patches in the following $T'$ consecutive frames will be masked; We repeat this process until $>$15\% of video patches are masked to perform MVM pre-training. The model cannot merely rely on similar neighboring visual cues but requires actual visual reasoning to recover a group of missing patterns. 

\vspace{0.5ex} \noindent\textbf{Attended Masking (AM).}
The conventional practice is to sample masked visual patches or textual tokens with the same probability over all visual and textual inputs. However, the important elements (\textit{e.g.}, visual patches containing the main object or content words) receive the same weight as the less relevant elements (\textit{e.g.}, scene background or stop words) in masked modeling. Attended Masking tries to put more weights on the more important elements based on the attention weights computed by Cross-modal Transformer (CT). A similar idea has been explored in~\cite{zellers2021merlot} for MLM. In this paper, we extend AM to both visual and textual modalities. We first keep the video-text inputs intact, feed them into CT to compute the attention weights, to decide which portions in video and text are more important. We then select the top 15\% of most-attended tokens to be masked in both video and text inputs to perform MVM and MLM.

\begin{table*}[t!]
\centering
    \subfloat[Text-to-video Retrieval (Finetuned).\label{table:retrieval}]{\tablestyle{3pt}{1.0}
    \def \w{15pt}
    \begin{tabular}{lcccc}
        \toprule
        ~ & \multicolumn{4}{c}{Text-to-Video Retrieval} \\
        \cmidrule{2-5}
        Method & MSRVTT & DiDeMo & YouCook2 & LSMDC \\
        \hline
        \hline
        \multicolumn{4}{l}{\textit{Models using Pre-extracted Features}}\\
        \hline
        HT100M~\cite{miech2019howto100m} & 14.9 / 40.2 / 52.8 & - & \textcolor{white}{0}8.2 / 24.5 / 35.3  & \textcolor{white}{0}7.1 / 19.6 / 27.9 \\
        \rowcolor{LightCyan}
        \demph{MMT~\cite{gabeur2020mmt}} & \demph{26.6 / 57.1 / 67.1} & \demph{-} & \demph{-} & \demph{12.9 / 29.9 / 40.1} \\
        \rowcolor{LightCyan}
        \demph{HERO~\cite{li2020hero}} & \demph{16.8 / 43.4 / 57.7} & \demph{-} & \demph{-} & \demph{-} \\
        \rowcolor{LightCyan}
        \demph{AVLnet~\cite{rouditchenko2021avlnet}} &  \demph{27.1 / 55.6 / 66.6} & \demph{-} & \demph{33.2 / 61.0 / 71.5} & \demph{17.0 / 38.0 / 48.6} \\
        Support-Set~\cite{patrick2021support-set} & 30.1 / 58.3 / 69.3 & - & - & -\\
        TACo~\cite{yang2021taco} & 28.4 / 57.8 / 71.2 & - & 29.6 / 59.7 / 72.7 & - \\
        VideoCLIP~\cite{xu2021video-clip} & 30.9 / 55.4 / 66.8 & - & 32.2 / 62.6 / 75.0 & - \\
        \hline
        \hline
        \multicolumn{4}{l}{\textit{Models with End-to-end Training}}\\
        \hline
        ClipBERT~\cite{lei2021clip-bert} & 22.0 / 46.8 / 59.9 & 20.4 / 48.0 / 60.8 & - & - \\
        Frozen~\cite{bain2021frozen} & 32.5 / 61.5 / 71.2 & 31.0 / 59.8 / 72.4 & - & 15.0 / 30.8 / 39.8 \\
        \demph{Clip4Clip~\cite{luo2021clip4clip}} &  \demph{42.1 / 71.9 / 81.4} & \demph{43.4 / 70.2 / 80.6} & \demph{-} & \demph{21.6 / 41.8 / 49.8}\\
        \modeltitle & \textbf{34.5} / \textbf{63.0} / \textbf{73.4} & \textbf{32.6} / \textbf{62.8} / \textbf{74.7} & \textbf{35.7} / \textbf{66.7} / \textbf{78.2} & \textbf{16.1} / \textbf{36.6} / \textbf{41.2} \\
        \bottomrule
    \end{tabular}
    }
    \hfill
    \subfloat[Text-to-video Retrieval (Zero-shot).\label{table:retrieval-zs}]{\tablestyle{3pt}{1.0}
    \def \w{15pt} 
    \begin{tabular}{lcc}
        \toprule
        ~ & \multicolumn{2}{c}{Zero-Shot Retrieval} \\
        \cmidrule{2-3}
        Method & MSRVTT & DiDeMo \\
        \hline
        \hline
        \multicolumn{3}{l}{\textit{Models using Pre-extracted Features}} \\
        \hline
        HT100M~\cite{miech2019howto100m} & \textcolor{white}{0}7.5 / 21.2 / 29.6 & - \\
        \rowcolor{LightCyan}
        \demph{MMT~\cite{gabeur2020mmt}} & \demph{\textcolor{LightCyan}{0}-\textcolor{LightCyan}{0} / \textcolor{LightCyan}{0}6.9 / \textcolor{LightCyan}{0}-\textcolor{LightCyan}{0}} & \demph{-} \\
        \rowcolor{LightCyan}
        \demph{AVLnet~\cite{rouditchenko2021avlnet}} &  \demph{19.6 / 40.8 / 50.7} & \demph{-} \\
        Support-Set~\cite{patrick2021support-set} &  12.7 / 27.5 / 36.2 & - \\
        TACo~\cite{yang2021taco} & \textcolor{LightCyan}{0}9.8 / 25.0 / 33.4 & - \\
        VideoCLIP~\cite{xu2021video-clip} & 10.4 / 22.2 / 30.0 & 16.6 / 46.9 / \textcolor{LightCyan}{0}-\textcolor{LightCyan}{0} \\
        \hline
        \hline
        \multicolumn{3}{l}{\textit{Models with End-to-end Training}} \\
        \hline
        MIL-NCE~\cite{miech2020end} & \textcolor{white}{0}9.9 / 24.0 / 32.4 & - \\
        \rowcolor{LightCyan}
        \demph{VATT~\cite{akbari2021vatt}} &  \demph{\textcolor{LightCyan}{0}-\textcolor{LightCyan}{0} / \textcolor{LightCyan}{00}-\textcolor{LightCyan}{0} / 29.7} & \demph{-} \\
        Frozen~\cite{bain2021frozen} & 24.7 / 46.2 / 57.2 & 21.1 / 46.0 / 56.2 \\
        \demph{CLIP~\cite{radford2021clip,luo2021clip4clip}} &  \demph{31.2 / 53.7 / 64.2} & \demph{-} \\
        \modeltitle & \textbf{25.9} / \textbf{49.5} / \textbf{59.7} & \textbf{23.5} / \textbf{49.8} / \textbf{59.8} \\
        \bottomrule
    \end{tabular}
    }
    \vspace{-2ex}
    \caption{Comparison with SOTA on 
    \textbf{text-to-video-retrieval} tasks under different settings: (a) pre-train then finetune and (b) pre-train then zero-shot evaluation. All resutls are reported on R@1 / R@5 / R@10. All models perform visual-text pre-training. Rows highlighted in blue use additional modalities such as sound and speech besides video frames.}
    \vspace{-2ex}
    \label{table:retrieval-all}
\end{table*}

\section{Experiments}
\subsection{Experimental Setup}
\vspace{0.5ex} \noindent\textbf{Pre-training Datasets.}
As mentioned in Sec.~\ref{sec:model}, \modelname is flexible in taking both video and image as inputs. Hence, We follow~\cite{bain2021frozen} to jointly pre-train our model on image-text and video-text data, which we briefly describe below. ($i$) \textbf{YT-Temporal-180M (YT-Temporal)}~\cite{zellers2021merlot} contains 6M YouTube videos with subtitle texts from Automatic Speech Recognition (ASR). Following~\cite{zellers2021merlot}, we divide a long video into several video segments, with an average length of 9.29 seconds. We treat every 4 consecutive segments with their ASR as a video clip, leading to 180M video-subtitle pairs. ($ii$) \textbf{WebVid-2.5M (WebVid)}~\cite{bain2021frozen} scrapes 2.5M video-text pairs from the web. 
Different from YT-Temporal, 
text data in WebVid describes the global video semantic. ($iii$) \textbf{ConceptualCaptions-3M (CC)}~\cite{sharma2018cc} consists of 3.3M image-text pairs harvested from the web. We compare the effects of different pre-training data on downstream tasks in Sec.~\ref{sec:pretraining-data}.

\vspace{0.5ex} \noindent\textbf{Downstream Tasks.}
We evaluate \modelname on both text-to-video retrieval and video question answering, across 12 downstream benchmarks. For text-to-video retrieval, we report performance of Recall at K (R@K) on MSRVTT~\cite{xu2016msrvtt}, DiDeMo~\cite{hendricks2017didemo}, YouCook2~\cite{zhou2018youcook2} and LSMDC~\cite{rohrbach2015lsmdc}. For video question answering, we consider 8 datasets in multiple-choice and open-ended settings: TGIF-Action, TGIF-Transition and TGIF-Frame~\cite{jang2017tgif-qa}, MSRVTT-MC~\cite{yu2018js-fusion}, MSRVTT-QA, MSVD-QA~\cite{xu2017msrvtt-qa}, LSMDC-MC and LSMDC-FiB ~\cite{torabi2016lsmdc-fib}. Accuracy is used as evaluation metric. More details are provided in Appendix~\ref{sec:downstream}.

\vspace{0.5ex} \noindent\textbf{Implementation Details.}
We initialize our Video Swin Transformer with VideoSwin-Base~\cite{liu2021video-swin}, pre-trained on Kinetics-400~\cite{kay2017kinetics}. Language Embedder and Cross-modal Transformer are initialized from pre-trained BERT-Base~\cite{su2020vlbert}. We train \modelname in a end-to-end manner for both pre-training and downstream finetuning.

During pre-training, we sparsely sample $T$ = 4 video frames and resize them into 224x224 to split into patches with $H$ = $W$ = 32. We use pre-trained DALL-E~\cite{ramesh2021dalle} as our dVAE to generate discrete visual tokens for MVM. For WebVid~\cite{bain2021frozen} and CC~\cite{sharma2018cc}, we perform VTM+MLM+MVM to pre-train on videos or images with the globally-aligned alt-text descriptions. We follow~\cite{zellers2021merlot} to concatenate all ASR descriptions for each middle frame as text input for YT-Temporal. VTM is performed for each pair of middle frame and its ASR text to learn the temporal reasoning over YT-Temporal video clips. Our implementation of \modelname is based on PyTorch~\cite{paszke2019pytorch}. We adopt AdamW~\cite{loshchilov2019adamw} as the optimizer with an initial learning rate of 2e-5, betas of (0.9, 0.98), and weight decay of 1e-3 for all pre-training experiments. \modelname follows a simple curriculum learning strategy, where we first pre-train on YT-Temporal with noisy ASR text for 5 epochs and then on WebVid+CC with alt-text descriptions for another 5 epochs. 

For all downstream tasks, we adopt the same video frame size (224x224) and patch size (32x32) but 5 sparse-sampled frames. Due to various data scales and domains, we use task-specific learning rates and training epochs based on the performance of the validation set for each downstream task.

\subsection{Comparison to Prior Arts} \label{sec:main_results}
\vspace{0.5ex} \noindent\textbf{Text-to-Video Retrieval.}
Table~\ref{table:retrieval} summarizes results on text-to-video retrieval. \modelname achieves significant gain over existing VidL pre-trained models across all text-to-video retrieval datasets considered. Specifically, \modelname surpasses most previous methods focus on modeling multi-modal fusion with pre-extracted video features. Notably, \modelname is still competitive even when compared with MMT~\cite{gabeur2020mmt}, HERO~\cite{li2020hero}, and AVLnet~\cite{rouditchenko2021avlnet} that use additional modalities, such as sound and speech besides video frames. 

For comparisons to end-to-end pre-trained models, \modelname outperforms ClipBERT~\cite{lei2021clip-bert} by $+10\%$ on R@1 on both MSRVTT and DiDeMo, even though \modelname uses even less frames (Ours: 5 frames vs. ClipBERT: 16 frames). These results highlight the deficiency of `imagifying` video representations. When compared with Frozen~\cite{bain2021frozen}, designed specifically for text-to-video retrieval tasks, \modelname can achieve notable performance improvements with $+2.0\%$, $+1.6\%$ and $+1.1\%$ on R@1 for MSRVTT, DiDeMo and LSMDC, respectively. We also include results from Clip4Clip~\cite{luo2021clip4clip} that leverages pre-trained CLIP~\cite{radford2021clip} on over 400M image-text data, which is a few magnitude larger than our pre-training data. \modelname closes the gap between previous end-to-end pre-trained models and Clip4Clip, and we believe pre-training \modelname with larger-scale data can further reduce the gap.

\vspace{0.5ex} \noindent\textbf{Zero-shot text-to-video retrieval.} 
We further conduct generalizability evaluation under the zero-shot setting on MSRVTT and DiDeMo in Table~\ref{table:retrieval-zs}. Similarly, \modelname achieves remarkable performance improvements over the existing methods by large margins. Specifically, we observe $+6\%$ gain on R@1 over previous models using pre-extracted video features and $+1.2-2\%$ on R@1 over end-to-end pre-trained models, excluding CLIP~\cite{radford2021clip,luo2021clip4clip}.

\begin{table}[t]
\centering
    \tablestyle{2pt}{1.0} 
    \def \w{20pt} 
    \resizebox{\linewidth}{!}{
    \begin{tabular}{lccccccccccc}
        \toprule
        ~ & \multicolumn{3}{c}{TGIF} & ~ & \multicolumn{2}{c}{MSRVTT} & ~ & \multicolumn{2}{c}{LSMDC} & ~ & MSVD \\
        \cmidrule{2-4} \cmidrule{6-7} \cmidrule{9-10} \cmidrule{12-12} Method & Action & Transition & Frame & ~ & MC & QA & ~ & MC & FiB & ~ & QA \\
        \midrule
        ClipBERT~\cite{lei2021clip-bert} & 82.8 & 87.8 & 60.3 & ~ & 88.2 & 37.4 & ~ & - & - & ~ & - \\
        JustAsk~\cite{yang2021just-ask} & - & - & - & ~ & - & 41.5 & ~ & - & - & ~ & 46.3 \\
        \textcolor{lightgray}{MERLOT}~\cite{zellers2021merlot} & \textcolor{lightgray}{94.0} & \textcolor{lightgray}{96.2} & \textcolor{lightgray}{69.5} & ~ & \textcolor{lightgray}{90.9} & \textcolor{lightgray}{43.1} & ~ & \textcolor{lightgray}{81.7} & \textcolor{lightgray}{52.9} & ~ & - \\
        \modeltitle & \textbf{92.5} & \textbf{95.7} & \textbf{68.9} & ~ & \textbf{91.9} & \textbf{43.9} & ~ & \textbf{82.8} & \textbf{53.7} & ~ & \textbf{47.9} \\
        \bottomrule
    \end{tabular}
    }
    \vspace{-2ex}
    \caption{Comparison with SOTA methods on \textbf{video question answering}. We gray out MERLOT due to its excessive computational cost (\textit{e.g.,} 30K TPU hours \textit{vs.} 2K GPU hours (ours) for pre-training and frame resolution 704 \textit{vs.} 224 for downstream tasks).}
    \vspace{-2ex}
    \label{table:qa}
\end{table}

\vspace{0.5ex} \noindent\textbf{Video Question Answering.}
We compare with prior arts on video question answering (QA) tasks in Table~\ref{table:qa}. \modelname surpasses ClipBERT~\cite{lei2021clip-bert} with significant performance gain of $+9.7\%$ on TGIF-Action, $+7.9\%$ on TGIF-Transition, $+8.6\%$ on TGIF-Frame, $+3.7\%$ on MSRVTT-MC and $+6.5\%$ on MSRVTT-QA. These results suggest the explicit temporal modeling introduced by our video transformer is essential for video QA tasks, and pre-training with image-text data alone may not be sufficient for VidL modeling. We provide more detailed discussions in Sec.~\ref{sec:video-encoding}.

Note that both JustAsk~\cite{yang2021just-ask} and MERLOT~\cite{zellers2021merlot} specifically focus on video QA. JustAsk automatically generates 69M video-question-answer triplets from narrated videos for training, which is hardly extendable to text-to-video retrieval tasks. 
MERLOT is pre-trained for 40 epochs and with extensive hyperparameter tuning on the frame resolution from 384x704 to 704x704 for downstream tasks. The overall pre-training of MERLOT takes 30,720 TPU hours on TPU v3. In contrast, we pre-train \modelname for 5 epochs, which results in 2,240 GPU hours on V100 GPUs. We also adopt a much lower frame resolution of 224x224. With a much lower computational cost, \modelname achieves around $+1.0\%$ performance gain over MERLOT on 4 video QA tasks for MSRVTT and LSMDC videos, while remains competitive on TGIF. We believe \modelname can further improve with larger frame resolution and longer pre-training epoch if computational resources permit.

\begin{table}[t]
\centering
    \tablestyle{3pt}{1.0} 
    \def \w{20pt}
    \resizebox{\linewidth}{!}{
    \begin{tabular}{lcccc}
        \toprule
        Video & TGIF- & TGIF- & MSRVTT- & DiDeMo- \\
        Encoding & Action & Transition & Retrieval & Retrieval \\
        \hline
        \hline
        \multicolumn{5}{l}{\textit{Random initialized visual encoder}} \\
        \hline
        Mean & 72.1 & 83.5 & \textcolor{white}{0}8.4 / 22.7 / 35.3 & \textcolor{white}{0}9.1 / 24.9 / 36.7 \\
        Concat & 72.9 & 83.7 & \textcolor{white}{0}9.0 / 23.5 / 35.5 & \textcolor{white}{0}9.4 / 25.8 / 38.1 \\
        VT & \textbf{73.6} & \textbf{84.6} & \textbf{\textcolor{white}{0}9.2} / \textbf{24.0} /  \textbf{35.8}&\textbf{10.3} / \textbf{30.1} /  \textbf{40.5} \\
        \hline
        \hline
        \multicolumn{5}{l}{\textit{ImageNet pre-trained visual encoder}} \\
        \hline
        Mean & 77.5 & 86.5 & \textcolor{white}{0}9.6 /  26.7 / 39.5 & \textcolor{white}{0}9.5 / 27.5 / 40.9 \\
        Concat & 78.0 & 87.0 & 10.4 / 30.5 / 42.0&10.6 / 30.8 / 42.9 \\
        VT & \textbf{79.6} & \textbf{87.8} & \textbf{11.8} / \textbf{32.3} / \textbf{44.6} & \textbf{12.0} / \textbf{32.4} / \textbf{43.5} \\
        \hline
        \hline
        \multicolumn{5}{l}{\textit{+ Video-text pre-training on WebVid}} \\
        \hline
        Mean & 80.3 & 88.7 & 20.8 / 44.9 / 58.1&17.9 / 43.5 / 51.3 \\
        Concat & 82.5 & 91.2 & 23.5 / 51.9 / 63.0 & 22.2 / 50.5 / 62.6 \\
        VT & \textbf{85.8} & \textbf{92.1} & \textbf{27.0} / \textbf{56.5} / \textbf{68.8} & \textbf{26.1} / \textbf{56.9} / \textbf{68.9} \\
        \bottomrule
    \end{tabular}
    }
    \vspace{-2ex}
    \caption{Impact of \textbf{different  temporal modeling methods over video inputs} under different settings: ($i$) random initialized visual encoder; ($ii$) ImageNet~\cite{krizhevsky2012imagenet} pre-trained visual encoder and  ($iii$) Adding video-text pre-training on WebVid~\cite{bain2021frozen}.}
    \vspace{-2ex}
    \label{table:frame-encoding}
\end{table}

\subsection{Analysis of \modeltitle}
We conduct ablation experiments on two video question answering datasets (TGIF-Action and TGIF-Transition) and two text-to-video retrieval datasets (MSRVTT and DiDeMo) to study the factors leading to \modelname's success. 

\vspace{0.5ex} \noindent\textbf{Impact of Temporal Video Modeling.} \label{sec:video-encoding}
To demonstrate the necessity of temporal modeling even under sparse sampling, we compare three variants for temporal modeling in Table~\ref{table:frame-encoding}.  ($i$) \textbf{Mean}: mean-pooling over independently computed frame features via ResNet-50~\cite{he2016resnet} as in ~\cite{lei2021clip-bert}; ($ii$) \textbf{Concat}: concatenation of the aforementioned frame features along the temporal dimension as in~\cite{zellers2021merlot}; ($iii$) \textbf{VT}: enforcing spatial-temporal modeling altogether on input video frame sequences via Video Swin Transformer in \modelname. The final video representations are then concatenated with corresponding text embeddings and fed in Cross-modal Transformer for downstream VidL modeling. We show results under different settings: random-initialized visual encoder, ImageNet-pretrained visual encoder, and with additional VidL pre-training on WebVid~\cite{bain2021frozen}.

VT consistently outperforms Mean and Concat over the 4 datasets across all settings. The loss of temporal information in naive mean pooling (Mean) result in worst performance among the three. Although Concat can preserve the temporal order of input video frames, it solely relies on Cross-modal Transformer to model both the temporal dynamics in video and the correspondence between visual and textual elements, brings unsatisfactory performance.

When taking a closer look into different pre-training settings, multimodal pre-training on WebVid significantly boosts the model performance, compared to unimodal pre-training of visual encoder on ImageNet. In addition, VT benefits more from VidL pre-training, leading to a bigger performance gap when compared to Mean or Concat. As the exposure to video data during pre-training utmostly enhances the learning of temporal dynamics.

\begin{table}[t]
\centering
    \tablestyle{3pt}{1.0} 
    \def \w{20pt} 
    \resizebox{\linewidth}{!}{
    \begin{tabular}{lcccc}
        \toprule
        Pre-training & TGIF- & TGIF- & MSRVTT- & DiDeMo- \\
        Task & Action & Transition & Retrieval & Retrieval \\
        \midrule
        None & 81.9 & 88.5 & 13.0 / 36.5 / 49.6 & 18.3 / 46.4 / 56.5 \\
        VTM+MLM & 85.4 & 91.6 & 24.4 / 54.4 / 68.1 & 25.8 / 54.2 / 67.0 \\
        \midrule
        + MCM & 85.0 & 91.6 & 26.0 / 56.0 / 68.4 & 25.8 / 55.9 / \underline{68.1} \\
        + MFM & \underline{85.5} & \underline{92.0} & 26.2 / 55.5 / 68.4 & 25.4 / 55.5 / 67.8 \\
        + MPM & 85.0 & 91.8 & \underline{26.6} / \underline{56.2} / \underline{68.4} & \underline{26.0} / \underline{56.5} / 68.0 \\
        + MVM & \textbf{85.8} & \textbf{92.1} & \textbf{27.0} / \textbf{56.5} / \textbf{68.8} & \textbf{26.1} / \textbf{56.9} / \textbf{68.9} \\
        \bottomrule
    \end{tabular}
    }
    \vspace{-2ex}
    \caption{Impact of \textbf{self-supervised pre-training on video inputs}. All pre-training are conducted on WebVid~\cite{bain2021frozen}.}
    \vspace{-2ex}
    \label{table:visual-pretraining}
\end{table}

\begin{figure}[t]
\centering
    \includegraphics[width=\linewidth]{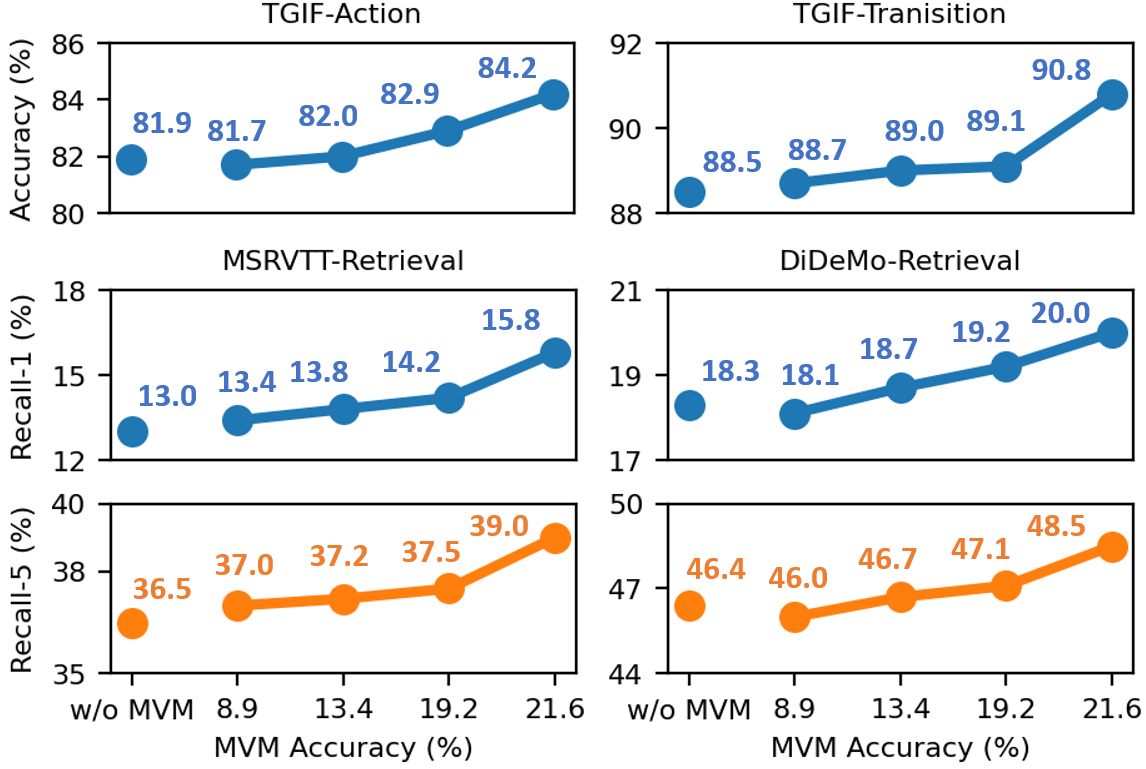}
    \vspace{-4ex}
    \caption{\textbf{MVM accuracy \textit{vs.} downstream performance}. We adopt only MVM during pre-training (using 0\% (w/o MVM), 10\% (8.9\% MVM accuracy), 20\% (13.4\%), 50\% (19.2\%), and 100\% (21.6\%) of YT-Temporal videos).}
    \label{fig:abl-mvm}
    \vspace{-2ex}
\end{figure}

\vspace{0.5ex} \noindent\textbf{Effectiveness of MVM.}
To demonstrate the effectiveness of MVM, we compare different variants of masked visual modeling when pre-trained on WebVid~\cite{bain2021frozen} in Table~\ref{table:visual-pretraining}. First, we establish two baselines: without pre-training (None) and pre-training with only Visual-Text Matching and Masked Language Modeling (VTM+MLM) following ~\cite{lei2021clip-bert,zellers2021merlot, bain2021frozen}. Then we augment VTM+MLM with different variants of masked visual modeling tasks. \textbf{Masked Classification Modeling (MCM)} mimics MRC in ~\cite{chen2020uniter} to predict the ImageNet~\cite{krizhevsky2012imagenet} category of the masked image patch from a pre-trained ResNet-50~\cite{he2016resnet}; \textbf{Masked Feature Modeling (MFM)}~\cite{li2020hero} distills the fixed frame features extracted from a pre-trained visual encoder. We use the output feature of the masked patches from the last CNN layer of a ImageNet pre-trained ResNet-50 and adopt linear regression as the training objective; \textbf{Masked Patch Modeling (MPM)}~\cite{li2020hero} distinguishes the correct masked visual patch from negative patches in the same batch with  Noise Contrastive Estimation loss~\cite{jozefowicz2016exploring}, similar to MFM-NCE in~\cite{li2020hero}.

Our results suggest that not all masked visual modeling methods bring consistent improvement. MCM and MPM give worse results on TGIF-Action over VTM+MLM; similar trends have been observed in~\cite{chen2020uniter,li2020hero}. MFM seems to favor QA tasks, and MPM benefits more on retrieval tasks. In contrast, MVM leads to the best performance on all tasks, as it recovers masked patches into a finite discrete set, making the learning of masked visual modeling easier.

We further investigate the relationship between MVM performance and downstream performance. We pre-train \modelname with MVM-only on 10\%, 20\%, 50\%, and 100\% of video scenes from YT-Temporal~\cite{zellers2021merlot}, discarding the corresponding text. As illustrated in Fig.~\ref{fig:abl-mvm}, such MVM pre-training on video inputs only can greatly lift the performance on all 4 datasets, even without text information. Moreover, better MVM performance also leads to better downstream performance. For example, with a 21.6\% MVM accuracy on 100\% YT-Temporal data, our model achieves  +2.3\% improvement on TGIF-Action and +2.5\% R@5 increase on MSRVTT. In summary, results in Table~\ref{table:visual-pretraining} and Fig.~\ref{fig:abl-mvm} suggest that MVM is vital in the success of \modelname, as it learns a better video representation to benefit downstream VidL tasks.

\begin{table}[t]
\centering
    \tablestyle{3pt}{1.0} 
    \def \w{20pt}
    \resizebox{\linewidth}{!}{
    \begin{tabular}{lcccc}
        \toprule
        \multirow{2}*{Method} & TGIF- & TGIF- & MSRVTT- & DiDeMo- \\
        ~ & Action & Transition & Retrieval & Retrieval \\
        \hline
        \hline
        \multicolumn{4}{l}{\textit{Without Pre-training}} \\
        \hline
        \modeltitle & 81.9 & 88.5 & 13.0 / 36.5 / 49.6 & 18.3 / 46.4 / 56.5 \\ 
        \hline
        \hline
        \multicolumn{4}{l}{\textit{Pre-training on COCO+VG}} \\
        \hline
        ClipBERT~\cite{lei2021clip-bert} & 82.8 & 87.8 & 22.0 / 46.8 / 59.9 & 20.4 / 48.0 / 60.8 \\
        \modeltitle & \textbf{84.8} & \textbf{90.2} & \textbf{23.5} / \textbf{50.5} / \textbf{63.9} & \textbf{22.8} / \textbf{51.2} / \textbf{62.0} \\
        \hline
        \hline
        \multicolumn{4}{l}{\textit{Pre-training on WebVid+CC}} \\
        \hline
        Frozen~\cite{bain2021frozen} & - & - & 31.0 / 59.5 / 70.5 & 31.0 / 59.8 / 72.4 \\
        \modeltitle & 87.1 & 93.6 & \textbf{34.2} / \textbf{63.5} / \textbf{73.6} & \textbf{32.9} / \textbf{63.0} / \textbf{74.5} \\
        \hline
        \hline
        \multicolumn{4}{l}{\textit{Pre-training on YTTemporal}} \\
        \hline
        \textcolor{lightgray}{MERLOT}~\cite{zellers2021merlot} & \textcolor{lightgray}{94.0} & \textcolor{lightgray}{96.2} & - & - \\
        \modeltitle & 91.0 & 94.7 & 25.4 / 54.3 / 64.6 & 26.7 / 56.4 / 64.6 \\
        \bottomrule
    \end{tabular}
    }
    \vspace{-2ex}
    \caption{Impact of using \textbf{different pre-training data}. We gray out MERLOT due to its excessive computational cost (\textit{e.g.,} 30K TPU hours vs. 2K GPU hours (ours) for pre-training and frame resolution 384x704 vs. 224x224 (ours) for downstream tasks).}
    \vspace{-2ex}
    \label{table:pretraining-data}
\end{table}

\vspace{0.5ex} \noindent\textbf{Impact of different pre-training data.} \label{sec:pretraining-data}
Table~\ref{table:pretraining-data} establishes a fair comparison to the recent SOTA methods, ClipBERT~\cite{lei2021clip-bert}, Frozen~\cite{bain2021frozen} and MERLOT~\cite{zellers2021merlot}, with the same pre-training data, respectively. We also compare the effects of different pre-training data on downstream tasks.

Under fair comparison, our model consistently outperforms ClipBERT and Frozen by large margins. When both pre-trained on COCO+VG~\cite{chen2015coco,krishna2017vg}, \modelname surpasses ClipBERT by $>$+2.0\% on Video QA tasks, and $>$+1.5\% on R@1 for retrieval tasks. Frozen adopts a two-stream architecture specifically designed for text-to-video retrieval applications. \modelname not only is applicable to video QA tasks but also achieves a gain of $>$+1.9\% on R@1 for retrieval tasks over Frozen, when both pre-trained on WebVid+CC~\cite{bain2021frozen,sharma2018cc}. On YT-Temporal~\cite{zellers2021merlot}, \modelname achieves competitive results with MERLOT on TGIF-Action and TGIF-Transition with a much lower training cost, as discussed in Sec.~\ref{sec:main_results}. We further examine the effect of different pre-training data on downstream tasks with \modelname. YT-Temporal is designed to promote video temporal reasoning and not surprisingly leads to the best QA result. However, the noisy ASR descriptions lead to smaller gains in retrieval tasks, with a similar performance to COCO+VG, but much worse than WebVid+CC with a smaller data size (5.5M vs. 180M). Therefore, we take advantage of both YT-Temporal and WebVid+CC as our final pre-training corpus, which leads to strong performance on both video QA and retrieval tasks as presented in Sec.~\ref{sec:main_results}.

\begin{figure}[t]
\centering
    \includegraphics[width=\linewidth]{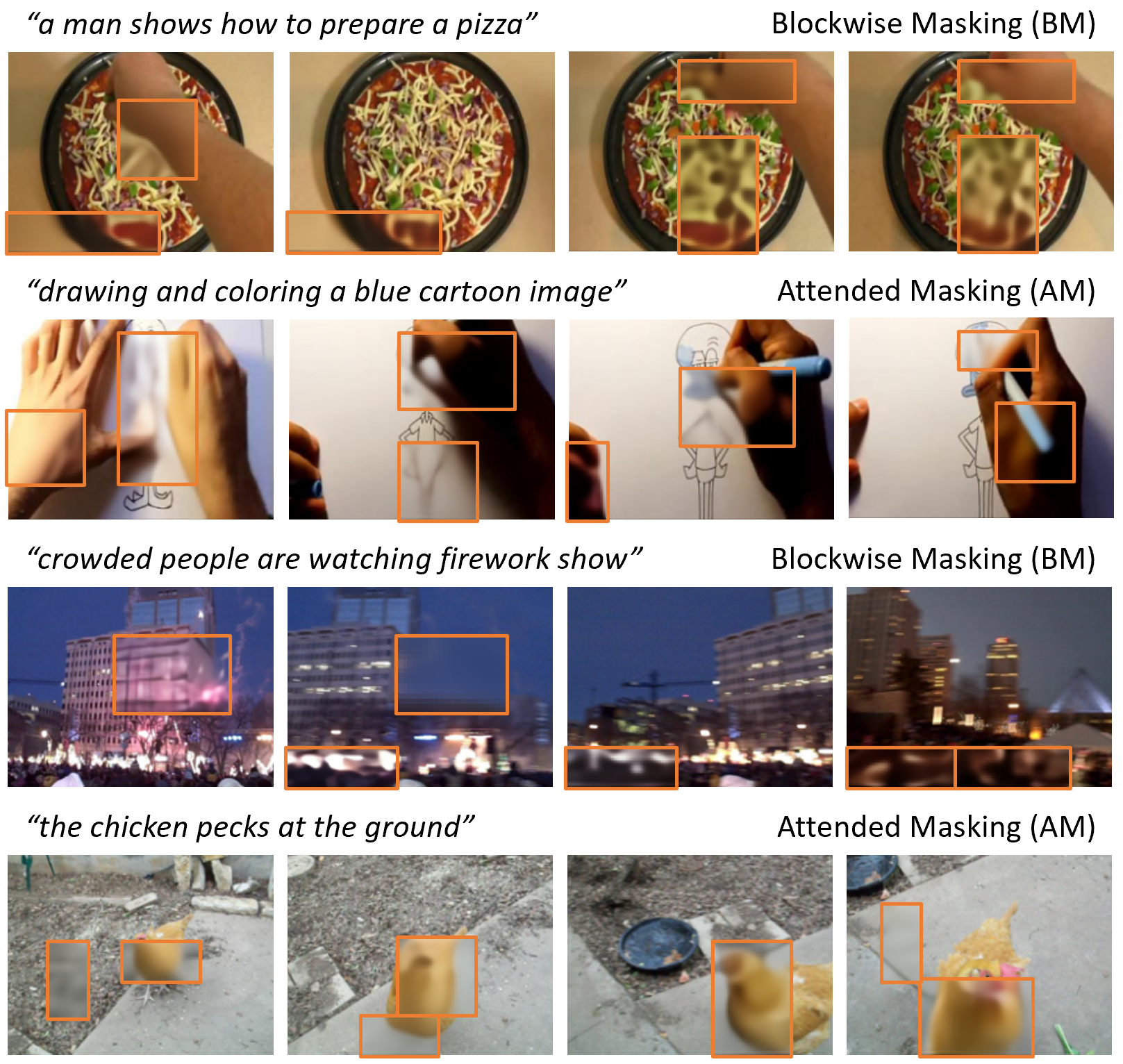}
    \vspace{-4ex}
    \caption{\textbf{Qualitative examples} of self-reconstruction (highlighted with orange bounding boxes) from predicted visual tokens during our Masked Visual-token Modeling (MVM).}
    \vspace{-2ex}
    \label{fig:qual}
\end{figure}

\vspace{0.5ex} \noindent\textbf{Qualitative Examples.}
Fig.~\ref{fig:qual} illustrates the qualitative examples of self-reconstruction from predicted visual tokens during MVM, under both Blockwise Masking (BM) and Attended Masking (AM).  As shown, BM masks blocks of video patches along with consecutive frames and AM masks the most-attended video patches based on text input (\textit{e.g.}, drawing with \textit{``hand''} and \textit{``cartoon image''} in the $2^\text{nd}$ row or \textit{``chicken''} and \textit{``ground''} in the $4^\text{th}$ row). \modelname improves visual reasoning through this video reconstruction during MVM, and the better video scene understanding further benefits downstream VidL tasks.

\section{Conclusion}
We present \modelname, a fully end-to-end VIdeO-LanguagE Transformer, which contains Video Swin Transformer to explicitly model the vital video temporal for video-language learning. We further enhance \modelname with a new pre-training task, Masked Visual-token Modeling (MVM), that learns video scene understanding through a mask-the-predict procedure with self-reconstructable visual tokens. Experiments on various text-to-video retrieval and video question answering tasks show that \modelname achieves SOTA (or competitive) performance. Comprehensive ablation studies demonstrate the necessity of temporal video modeling and the effectiveness of MVM over previous MRM/MFM for video-language reasoning under different pre-training settings. 

\appendix 

\section{Experimental Setup of Downstream Tasks}\label{sec:downstream}
We evaluate our pre-trained \modelname on text-to-video retrieval and video question answering tasks across 12 downstream datasets. For text-to-video retrieval, we report model performance on  MSRVTT~\cite{xu2016msrvtt}, DiDeMo~\cite{hendricks2017didemo}, YouCook2~\cite{zhou2018youcook2}, and LSMDC~\cite{rohrbach2015lsmdc} and use Recall at K (R@K) as the evaluation metric. For video question answering, we consider datasets in both multiple-choice and open-ended settings, including TGIF-Action, TGIF-Transition, TGIF-Frame~\cite{jang2017tgif-qa}, MSRVTT-MC, MSRVTT-QA, MSVD-QA~\cite{xu2017msrvtt-qa}, LSMDC-MC and LSMDC-FiB~\cite{torabi2016lsmdc-fib}. We evaluate our models using accuracy. 

We follow the standard training/validation/testing splits of the original datasets. If not otherwise stated, we sparsely sample $T$ = 5  video frames and adopt video frame size 224 with patch size 32. We use AdamW~\cite{loshchilov2019adamw} to fine-tune \modelname for each downstream task with an initial learning rate of 1.2e-5, betas of (0.9, 0.98), and weight decay of 1e-3. All finetuning experiments are conducted on Microsoft Azure~\cite{msft-azure} with 8 Nvidia V100 GPUs (32GB VRAM). 

\subsection{Text-To-Video Retrieval}
For text-to-video retrieval, similar to visual-text matching (VTM) during pre-training, we treat corresponding video-text pairs as positives and all other pairwise combinations as negatives. We adopt a fully-connected (FC) layer (FC$^\text{T2V}$) over the global VidL representation $h^\text{c}$ of the \texttt{[CLS]} token to perform binary classification:
\begin{equation}
\begin{split}
    b_\text{pos}&= \text{FC}^\text{T2V}(h^\text{c}_\text{pos}), b_\text{neg}= \text{FC}^\text{T2V}(h^\text{c}_\text{neg}), \\
    \mathcal{L}_\text{T2V} &= - \mathbb{E} [ \log(b_\text{pos})+\log(1-b_\text{neg})],
\end{split}
\end{equation}
where $h^\text{c}_\text{pos}$ or $h^\text{c}_\text{neg}$ is $h^\text{c}$ of positive or negative pairs. In particular, we use pre-trained FC$^\text{VTM}$ for zero-shot text-to-video retrieval and to  initialize FC$^\text{T2V}$ for further fine-tuning on each downstream text-to-video retrieval task.

\vspace{0.5ex} \noindent\textbf{MSRVTT~\cite{xu2016msrvtt}}
contains 10K YouTube videos with 200K human annotations. For fair comparison~\cite{bain2021frozen,lei2021clip-bert}, we train on 9K training+validation splits and evaluate on the 1K-A testing split. We adopt batch size 56 and train for 20 epochs. 

\vspace{0.5ex} \noindent\textbf{DiDeMo~\cite{hendricks2017didemo}}
consists of 10K videos annotated with 40K sentences from Flickr. Following~\cite{bain2021frozen,lei2021clip-bert}, we concatenate all sentences from the same video into a paragraph and perform paragraph-to-video retrieval for DiDeMo. We adopt batch size 48 and train for 20 epochs. 

\vspace{0.5ex} \noindent\textbf{YouCook2~\cite{zhou2018youcook2}}
contains 14K video clips from 2K cooking videos and 89 recipes. Each clip is annotated with one sentence. We follow~\cite{miech2019howto100m,xu2021video-clip} to report retrieval performance on the entire validation clips. We adopt batch size 56 and train for 40 epochs.

\vspace{0.5ex} \noindent\textbf{LSMDC~\cite{rohrbach2015lsmdc}}
is built upon 118K video clips from 202 movies. Each clip has a caption from movie scripts or descriptive video services. Following~\cite{bain2021frozen,miech2019howto100m}, we evaluate on 1K testing clips that disjoint from the training+validation splits. We adopt batch size 56 and train for 40 epochs.  

\begin{table}[t]
\centering
    \tablestyle{3pt}{1.0} 
    \def \w{20pt}
    \begin{tabular}{ccc}
        \toprule
        VideoQA & Task & \#Option \\
        \midrule
        \multirow{4}{*}{\shortstack{Multiple-\\Choice}} & TGIF-Action~\cite{jang2017tgif-qa} & 5 \\
        ~ & TGIF-Transition~\cite{jang2017tgif-qa} & 5 \\
        ~ & MSRVTT-MC~\cite{xu2017msrvtt-qa} & 5 \\
        ~ & LSMDC-MC~\cite{rohrbach2015lsmdc} & 5 \\
        \midrule
        \multirow{4}{*}{\shortstack{Open-\\Ended}} & TGIF-Frame~\cite{jang2017tgif-qa} & 1,540 \\
        ~ & MSRVTT-QA~\cite{xu2017msrvtt-qa} & 1,500 \\
        ~ & MSVD-QA~\cite{chen2011msvd-qa} & 1,000 \\
        ~ & LSMDC-FiB~\cite{torabi2016lsmdc-fib} & 908 \\
        \bottomrule
    \end{tabular}
    \vspace{-1ex}
    \caption{Summary of \textbf{video question answering} tasks.}
    \vspace{-2ex}
    \label{table:vqa}
\end{table}

\subsection{Video Question Answering}
We test our model on video question answering (QA) tasks in both multiple-choice and open-ended settings, as summarized in Table~\ref{table:vqa}. For multiple-choice QA tasks, we concatenate question with each answer option and add a separating blank token to form the input text (\texttt{Q+[SEP]+A}). We adopt a FC layer upon $h^\text{c}$ to predict the model confidence on each answer option.  Cross-entropy loss is used to train a classifier over all answer options for each video-question pair. For open-ended QA tasks, we follow the common practice to convert it to a classification task with a finite set of answer classes. We build a specific answer vocabulary that can cover most common answers in the training split of each dataset. Similarly, our model predicts the answer to a given question over all answer vocabulary through a FC layer upon $h^\text{c}$. 

\vspace{0.5ex} \noindent\textbf{TGIF-Action, TGIF-Transition, and TGIF-Frame~\cite{jang2017tgif-qa}}
require spatial-temporal reasoning to answer questions regarding GIF videos in TGIF-QA~\cite{jang2017tgif-qa} Specifically, we aim to test our model along three dimensions: ($i$) \textbf{Action}: to recognize the repeated action; ($ii$) \textbf{Transition}: to identify the transition between the before and after states; ($iii$) \textbf{Frame}: to answer questions about a specific frame from the GIF video. Among them, TGIF-Action and TGIF-Transition are collected under multiple-choice setting, and TGIF-Frame is an open-ended video QA task with free-form answers. In our implementation, we select 1,540 most common answers as answer candidates for TGIF-frame. We adopt batch size 48 and train for 20 epochs.

\vspace{0.5ex} \noindent\textbf{MSRVTT-MC and MSRVTT-QA~\cite{xu2017msrvtt-qa}}
are created based on videos and captions in MSRVTT~\cite{xu2016msrvtt}. MSRVTT-MC  is a multiple-choice task with videos as questions, and captions as answers. Each video contains 5 captions, with only one positive match. MSRVTT-QA contains 243K open-ended questions over 10K videos. We select 1,500 most common answers as the answer candidates. We adopt batch size 48 and training epochs 20  for both datasets.

\vspace{0.5ex} \noindent\textbf{MSVD-QA~\cite{xu2017msrvtt-qa}}
consists of 47K open-ended questions over 2K videos, based on video-caption pairs from MSVD~\cite{chen2011msvd-qa}. We use 1,000 most common answers as the answer vocabulary. We adopt batch size 80 and train for 40 epochs.

\vspace{0.5ex} \noindent\textbf{LSMDC-MC and LSMDC-FiB~\cite{torabi2016lsmdc-fib}} are built from LSMDC dataset~\cite{rohrbach2015lsmdc}. Similar to MSRVTT-MC, LSMDC-MC requires the model to select the only positive caption that describes the video from 5 caption candidates and formulates it as a multiple-choice QA task. LSMDC-FiB replaces a word in the question sentence with the \texttt{[BLANK]} token, and the model is to recover the missing word.  We regard LSMDC-FiB as an open-ended Video QA task. In particular, we use a FC layer over the joint VidL representation $h$ of the \texttt{[BLANK]} token to predict from 908 answer candidates. We adopt batch size 80 and train for 40 epochs.

\begin{table}[t]
\centering
    \tablestyle{3pt}{1.1} 
    \def \w{20pt} 
    \resizebox{\linewidth}{!}{
    \begin{tabular}{lcccc}
        \toprule
        Masking & TGIF- & TGIF- & MSRVTT- & DiDeMo- \\
        Strategy & Action & Transition & Retrieval & Retrieval \\
        \hline
        \hline
        \multicolumn{5}{l}{\textit{Without pre-training}}\\
        \hline
        None & 81.9 & 88.5 & 13.0 / 36.5 / 49.6 & 18.3 / 46.4 / 56.5 \\
        \hline
        \hline
        \multicolumn{5}{l}{\textit{Pre-train on WebVid~\cite{bain2021frozen} with VTM+MLM+MVM}} \\
        \hline
        Random & 83.7 & 90.8 & 24.3 / 54.8 / 66.7 & 24.2 / 53.5 / 67.6 \\
        BM & 85.4 & \underline{91.8} & \underline{27.0} / 56.2 / 68.6 & 25.8 / \underline{56.8} / \underline{68.8} \\
        AM & \underline{85.5} & 91.6 & 26.8 / \underline{56.5} / \underline{68.7} & \underline{26.0} / 56.8 / 68.6 \\
        BM+AM & \textbf{85.8} & \textbf{92.1} & \textbf{27.0} / \textbf{56.5} / \textbf{68.8} & \textbf{26.1} / \textbf{56.9} / \textbf{68.9} \\
        \bottomrule
    \end{tabular}
    }
    \vspace{-1ex}
    \caption{Impact of \textbf{masking strategy} in MVM and MLM. 
    }
    \vspace{-2ex}
    \label{table:bm-am}
\end{table}

\section{Impact of Masking Strategy}
To amplify the effectiveness of our MLM and MVM, we introduce Blockwise Masking (BM) and Attended Masking (AM) in Sec.~\ref{sec:masking}. Table~\ref{table:bm-am} compares different masking strategies when pre-trained on WebVid~\cite{bain2021frozen}. Specifically, we compare 4 masking strategies, random masking, BM only, AM only and BM+AM. Although improving from the non-pretrained baseline, random masking results in the least performance improvement on both video QA and retrieval tasks. In contrast, BM or AM alone brings more significant performance improvements on both tasks, while seems to benefit different tasks (\textit{e.g.,} 91.8\% on TGIF-Transition and 27.0\% R@1 on MSRVTT-Retrieval for BM, and 85.5\% on TGIF-Action and 26.0\% R@1 on DideMo-Retrieval for AM). Finally, by leveraging both BM and AM (BM+AM), we lead to the best performance among the four.

Unlike random masking, BM cuts down the spurious success in MVM evaluation through neighboring patches that are visually similar to the masked patches, and AM puts more masking weights on more important video-text elements based on the attention pattern from our Cross-modal Transformer. These results demonstrate that both BM and AM contribute to the success of \modelname.

\section{Extending \modeltitle to Image Question Answering Task}
In this section, we show that \modelname is also extendable to image question answering task by evaluating it on VCR~\cite{zellers2019vcr}, which requires commonsense reasoning about the image content. We follow MERLOT~\cite{zellers2021merlot} to draw colored highlights around the referenced entity (\textit{e.g.,} \texttt{[PERSON-1]} and \texttt{[CHAIR-2]}) in the given image and report performance on the  multiple-choice Q→A subtask. To finetune our model,  we concatenate the question and each answer choice from the 4 possible answer candidates. Similarly, a FC layer upon the global cross-modal representation $h^\text{c}$ of the \texttt{[CLS]} token is adopted to predict the answer and cross-entropy loss is used to supervise the model training. We adopt batch size 48 and train for 20 epochs. 
 
 \begin{table}[t]
\centering
    \tablestyle{3pt}{1.0} 
    \def \w{20pt}
    \begin{tabular}{ccc}
        \toprule
        Method & Frame Size & VCR \\
        \midrule
        MERLOT~\cite{zellers2021merlot} & 384x704 & 75.1 \\
        \midrule
        \modeltitle & 224x224 & 74.9 \\
        \modeltitle & 384x384 & \textbf{76.3} \\
        \bottomrule
    \end{tabular}
    \vspace{-1ex}
    \caption{Comparison with MERLOT~\cite{zellers2021merlot} under the same pre-training epoch on VCR~\cite{zellers2019vcr}. The pre-training are conducted on YT-Temporal~\cite{zellers2021merlot} for 5 epochs.}
    \vspace{-2ex}
    \label{table:vcr}
\end{table}
The results are shown in Table~\ref{table:vcr}. For a fair comparison, both \modelname and MERLOT are pre-trained on YT-Temporal~\cite{zellers2021merlot} for 5 epochs. Note that MERLOT adopts a input image resolution of 384x704.  With input image size of 224x224, our \modelname achieves comparable performance  as MERLOT (74.9\% \textit{vs.} 75.1\%). When increasing the input image resolution to 384x384, though still smaller than the input image size in MERLOT, \modelname can achieve a superior performance with an absolute gain of +1.2\% over MERLOT.  

As mentioned in Sec.~\ref{sec:main_results}, the full MERLOT pre-training requires excessive computation power (30K TPU hours), while \modelname, only pre-trained for 5 epochs (2K GPU hours), can achieve competitive performance on video-language downstream tasks with lower input resolution (ours: 224x224, MERLOT:384x704). When comparing the results from \modelname with different input resolutions in Table~\ref{table:vcr}, we observe that higher input resolution results in better downstream performance. It is also worth noting that longer pre-training leads to monotonic performance improvement, as shown in ~\cite{zellers2021merlot}. Hence we believe \modelname can further improve with higher frame resolution and more pre-training epochs if computational resources permit.

\section{Qualitative Examples of Zero-shot Text-to-Video Retrieval}
\begin{figure*}[t]
\centering
    \includegraphics[width=\linewidth]{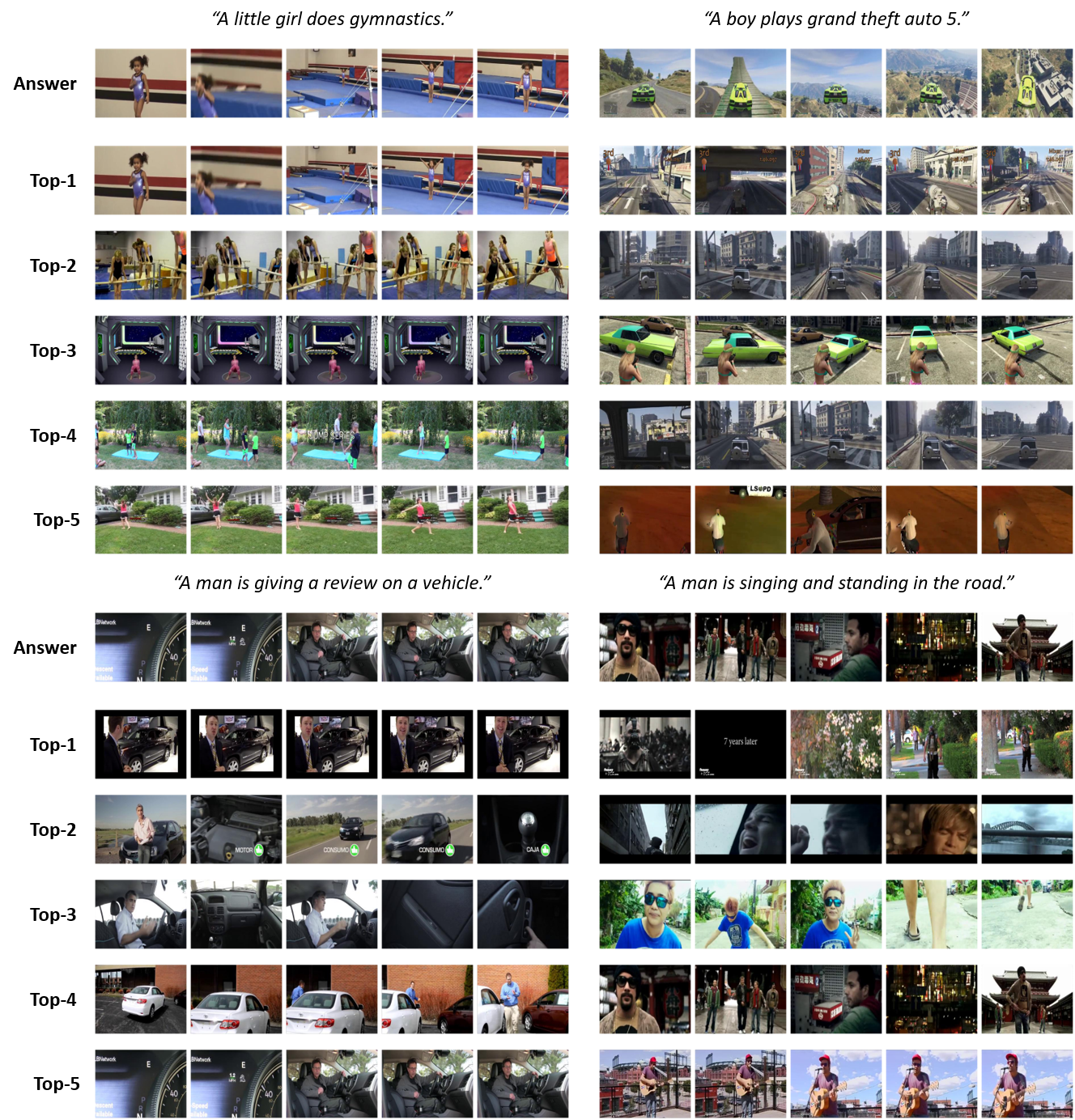}
    \vspace{-2ex}
    \caption{Qualitative examples of \textbf{zero-shot text-to-video retrieval} on MSRVTT~\cite{xu2016msrvtt}.}
    \vspace{-2ex}
    \label{fig:zs-msrvtt}
\end{figure*}

\begin{figure*}[t]
\centering
    \includegraphics[width=\linewidth]{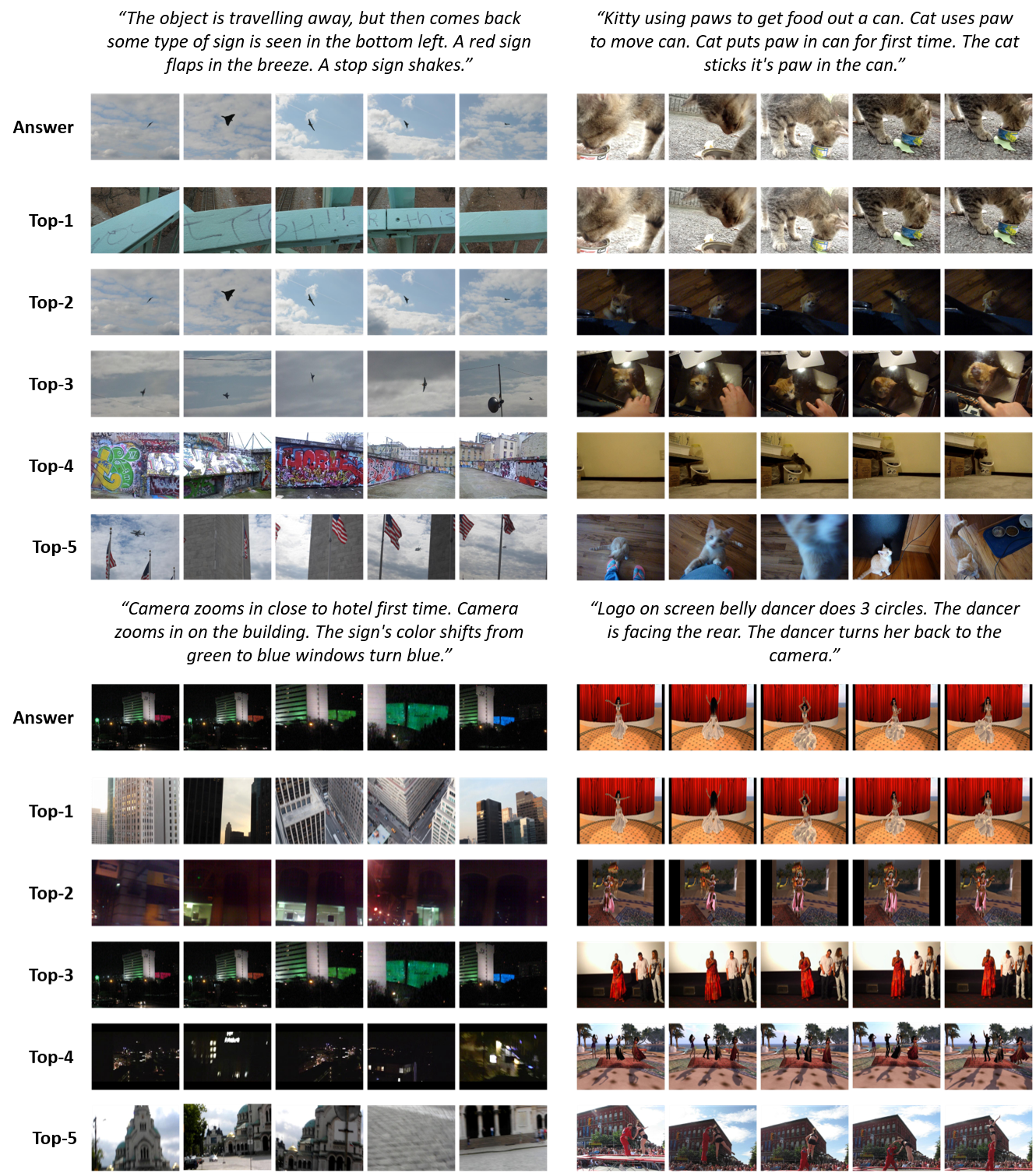}
    \vspace{-2ex}
    \caption{Qualitative examples of \textbf{zero-shot text-to-video retrieval} on DiDeMo~\cite{hendricks2017didemo}.}
    \vspace{-2ex}
    \label{fig:zs-didemo}
\end{figure*}

\begin{figure*}[t]
\centering
    \includegraphics[width=\linewidth]{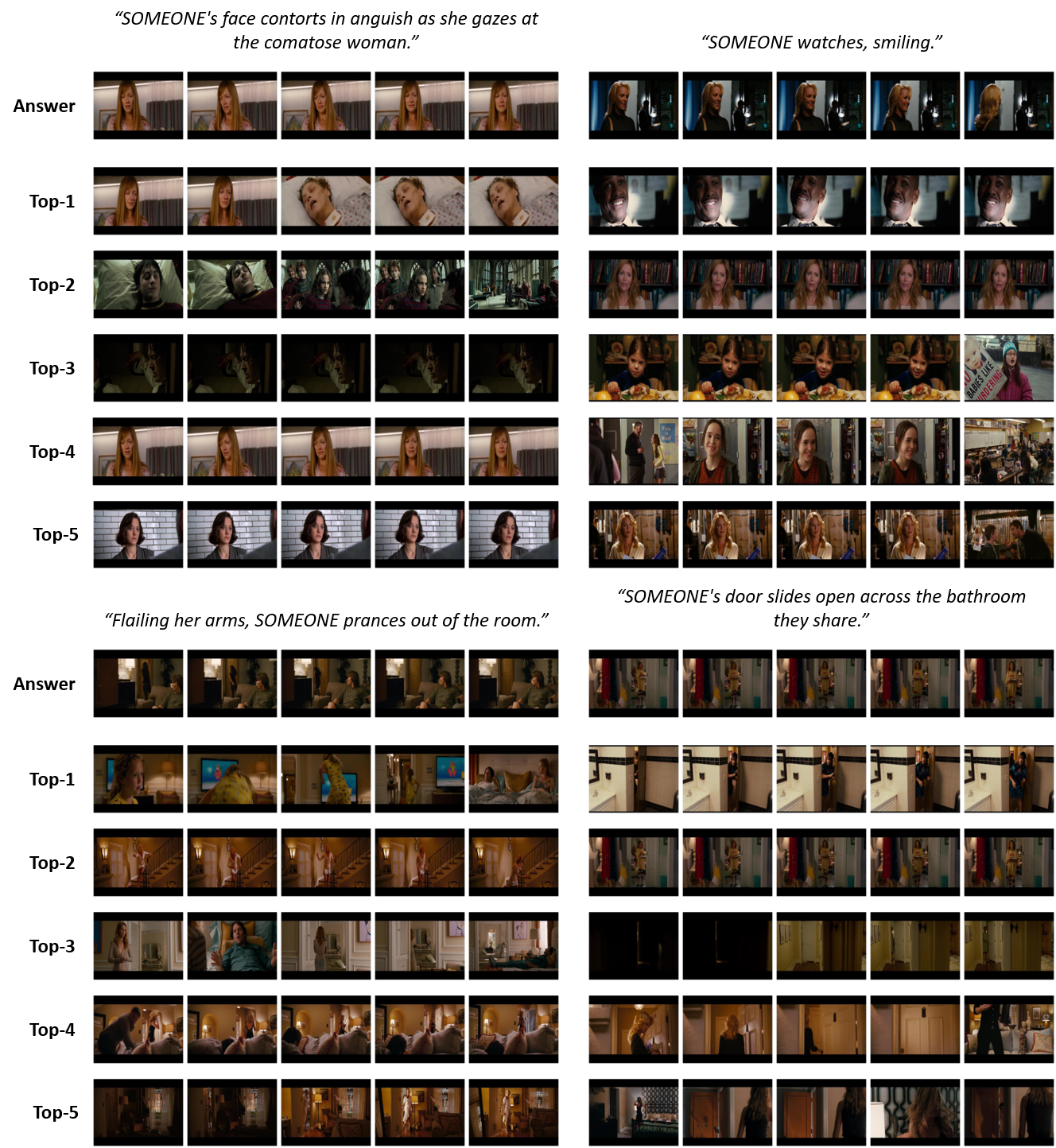}
    \vspace{-2ex}
    \caption{Qualitative examples of \textbf{zero-shot text-to-video retrieval} on LSMDC~\cite{rohrbach2015lsmdc}.}
    \vspace{-2ex}
    \label{fig:zs-lsmdc}
\end{figure*}

\begin{figure*}[t]
\centering
    \includegraphics[width=\linewidth]{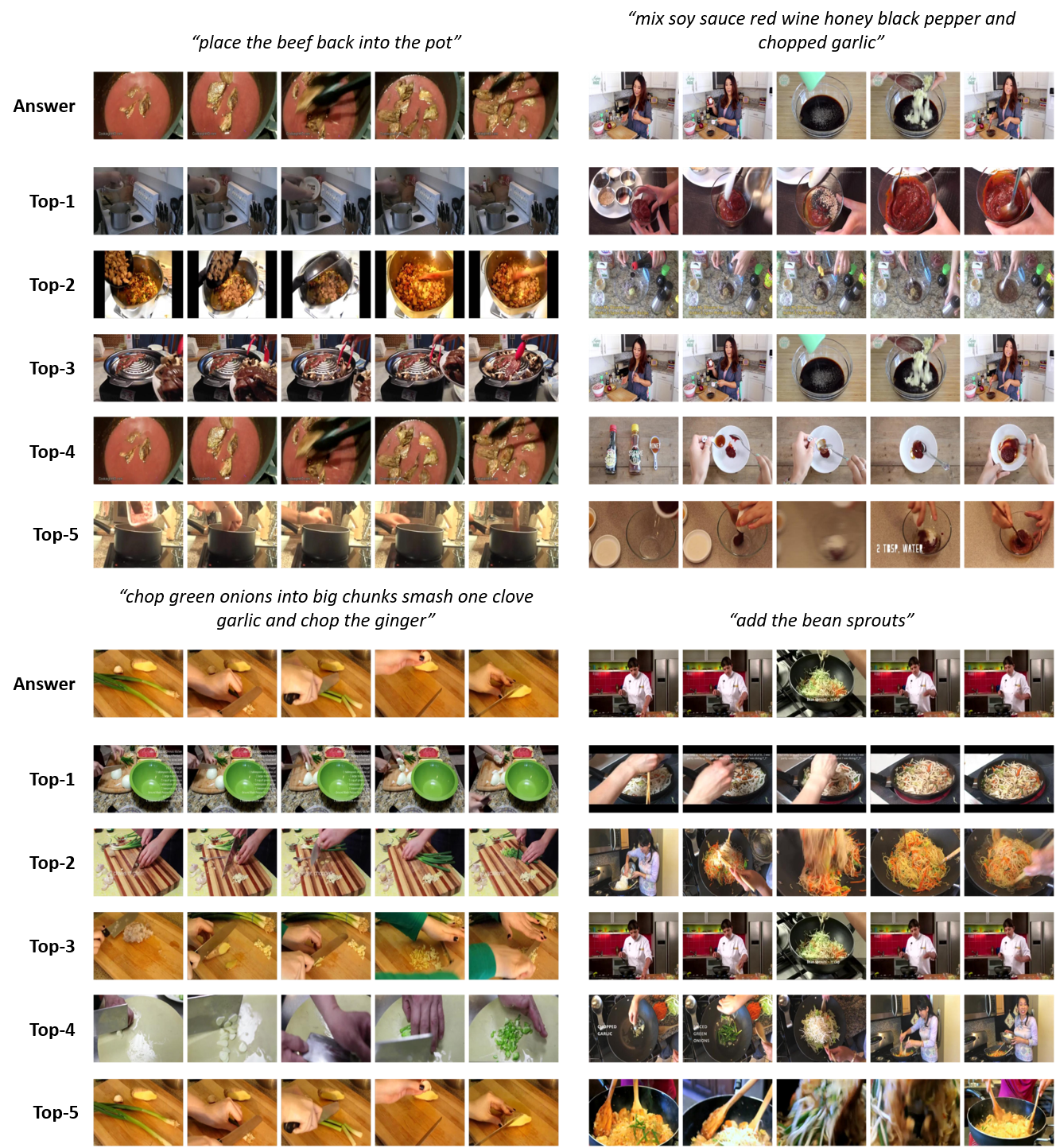}
    \vspace{-2ex}
    \caption{Qualitative examples of \textbf{zero-shot text-to-video retrieval} on YouCook2~\cite{zhou2018youcook2}.}
    \vspace{-2ex}
    \label{fig:zs-youcook}
\end{figure*}

We visualize some qualitative examples of zero-shot text-to-video retrieval in Fig.~\ref{fig:zs-msrvtt}-\ref{fig:zs-youcook} on MSRVTT~\cite{xu2016msrvtt}, DiDeMo~\cite{hendricks2017didemo}, LSMDC~\cite{rohrbach2015lsmdc}, and YouCook2~\cite{zhou2018youcook2}, respectively. These examples show that pre-training on large-scale visual-text data (YT-Temporal~\cite{zellers2021merlot}, WebVid~\cite{bain2021frozen}, and CC~\cite{sharma2018cc}) enables \modelname to learn cross-modal alignment to  perform text-to-video retrieval in a zero-shot scenario. 
For MSRVTT (Fig.~\ref{fig:zs-msrvtt}), \textit{``grand theft auto 5''} (a video game) is not a commonly seen phrase, but we can still retrieve relevant video clips depicting the video game. For paragraph-to-video retrieval in DiDeMo (Fig.~\ref{fig:zs-didemo}), the textual query is a concatenation of multiple sentences, much longer than the input text during pre-training. Surprisingly, \modelname can still retrieve videos that contain relevant semantics mentioned in the textual query. For instance, the top-2/3/5 of the retrieved videos on the upper left of Fig.~\ref{fig:zs-didemo} correspond to the textual cues, such as \textit{``traveling away, comes back, red signs flaps''}, Moreover, visualizations of zero-shot text-to-video retrieval on LSMDC (Fig.~\ref{fig:zs-lsmdc}) and YouCook2 (Fig.~\ref{fig:zs-youcook}) show that \modelname is generalizable to more specific video domains, such as movie or cooking videos. 

\section{Limitation and Broader Impact}
The broader impact of this paper falls in applications of video-language (VidL) reasoning, including video question answering and text-to-video retrieval. Our end-to-end VIdeO-LanguagE Transformer (\modelname) has the potential to be applied to various VidL tasks, such as video captioning and video grounded dialogue, which is worth exploration in future study. In addition, the newly introduced Masked Visual-token Modeling can further improve the performance when scaling up the pre-training to even larger-scale visual-text data. There are also several potential limitations of \modelname that would make for promising avenues of future work, including: 1) extending \modelname to model the full-length videos with densely sampled frames for downstream VidL tasks like TGIF-Count~\cite{jang2017tgif-qa}; and 2) exploring extra input signals from videos, such as audio, into the \modelname framework for better performance.

We do not anticipate major ethical issues in our work. As a data-driven system, the self-supervised method is sensitive to the distribution of the pre-training data. Therefore, we consider diverse types of video, including VLOGs, instructional videos, short-duration GIFs, and even static images across YT-Temporal~\cite{zellers2021merlot}, WebVid~\cite{bain2021frozen}, and CC~\cite{sharma2018cc}. The accompanying textual inputs used to train our model are from various sources, such as alt-text descriptions, human annotations, and ASR outputs, at different levels, including temporally-specified and globally-aligned descriptions. We conduct a comprehensive downstream evaluation over 12 VidL tasks, trying to mitigate the bias of our learned cross-modal representation for better VidL reasoning.

\clearpage

\clearpage

\clearpage

\clearpage
 
\clearpage

{\small
\bibliographystyle{ieee_fullname}
\bibliography{egbib}
}

\end{document}